\documentclass[11pt]{article}

\usepackage[preprint]{acl}

\usepackage{times}
\usepackage{latexsym}

\usepackage[T1]{fontenc}

\usepackage[utf8]{inputenc}

\usepackage{microtype}

\usepackage{inconsolata}

\usepackage{graphicx}
\usepackage{amsmath}
\usepackage{amssymb} 
\usepackage{booktabs}
\usepackage{tabularx}
\usepackage{mathtools}
\usepackage{arydshln} 
\usepackage{multirow} 
\usepackage{stfloats}
\usepackage[figuresright]{rotating}
\usepackage{float}
\usepackage{url}
\usepackage{listings}
\usepackage{xcolor}
\usepackage{CJKutf8}
\newcommand{\zh}[1]{\begin{CJK*}{UTF8}{gbsn}#1\end{CJK*}}
\newcommand{\rev}[1]{{\color{blue} #1}}
\renewcommand{\rev}[1]{{#1}}

\title{Rewrite to Translate, Translate to Reward: Reinforcement Learning for Source Rewriting in Machine Translation}

\author{Boxuan Lyu$^{\text{1}}$, Haiyue Song$^{\text{2}}$, Zhi Qu$^{\text{3}}$,
 Hidetaka Kamigaito$^{\text{3}}$, Kotaro Funakoshi$^{\text{1}}$, and Manabu Okumura$^{\text{1}}$ \vspace{1mm}\\ 
  $^{\text{1}}$Institute of Science Tokyo, $^{\text{2}}$Preferred Networks, Inc., \\$^{\text{3}}$Nara Institute of Science and Technology\\
  \texttt{\url{{lyu,funakoshi,oku}@lr.first.iir.isct.ac.jp}}\\ \texttt{\url{haiyuesong@preferred.jp}}\\ \texttt{\url{{qu.zhi.pv5,kamigaito.h}@is.naist.jp}}}

\begin{document}
\maketitle
\begin{abstract}
Prior work has explored prompting large language models (LLMs) to rewrite source text before translation, with the goal of improving machine translation (MT) quality.
However, we find that such prompt-based rewriting can degrade translation quality rather than enhance it, particularly when smaller LLMs, such as 4B-parameter models, are used. 
We argue that this limitation stems from the difficulty of controlling rewriting behavior through natural-language prompts alone: a rewrite is useful only if it leads to a better downstream translation, yet existing prompt-based methods do not explicitly optimize for this signal. 
To address this issue, we propose \textbf{RLSR} (\textbf{R}einforcement \textbf{L}earning for \textbf{S}ource \textbf{R}ewriting), a reinforcement learning framework that trains the rewriting model with a reward derived from the downstream translation-quality improvement produced by each rewrite. 
Experiments across six MT models and 16 language pairs show that our 4B RLSR-trained rewriting models significantly outperform both the no-rewriting baseline and same-scale prompt-based rewriting baselines, while remaining competitive with baselines that use a 235B LLM.
Our models and code are available at: \url{https://github.com/vlaks425/MT-RLSR}
\end{abstract}

\section{Introduction}

Source rewriting, sometimes referred to as pre-editing, modifies the source text before it is passed to a machine translation (MT) model, with the aim of improving translation quality. 
When the internal workings of an MT model are inaccessible, source rewriting provides a practical remedy \cite{hpe1, hpe2, hpe3, fujita21, rewriting}. 
Recent work has explored prompting large language models (LLMs) directly to perform such rewriting \cite{rewriting_qe,rewriting}.
\begin{figure}
    \centering
    \includegraphics[width=1\linewidth]{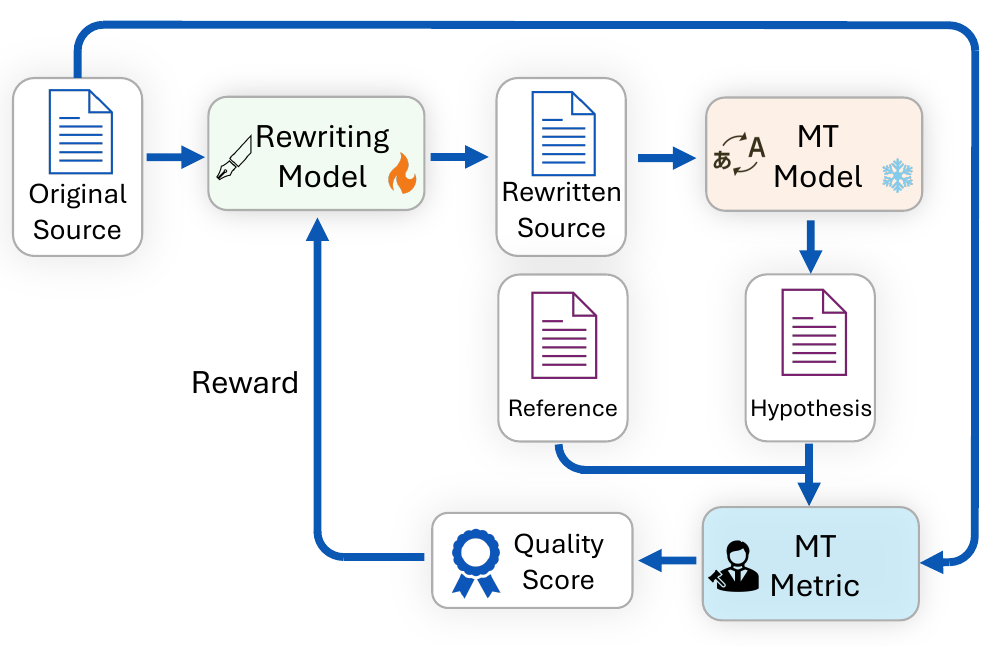}
    \caption{Overview of RLSR. The rewriting model generates a rewritten source from the original source. A fixed downstream MT model translates the rewritten source, and an MT metric evaluates the translation. The improvement over translating the original source is used as the reward for optimizing the rewriting model.}
    \label{fig:overview}
\end{figure}
However, we find that prompt-based rewriting is unreliable: rather than consistently improving translation quality, it sometimes degrades it, especially for smaller LLMs, such as 4B-parameter models.
We argue that this unreliability is inherent to controlling rewriting through natural-language prompts alone.
A prompt can express only an intended behavior, such as ``simplify the text'' or ``make it easier to translate,'' whereas the practical utility of a rewrite depends on whether it leads the downstream MT model to produce a better translation.
Prompt-based methods neither observe this utility nor optimize for it; they instead rely on the assumption that following the instruction will improve the translation.
This raises a central question: \textit{how can we obtain a rewriting model that reliably improves translation quality, without relying on prompts that steer the rewrite indirectly?}

To address this question, we optimize the rewriting model \emph{directly} for the objective of interest, rather than using prompts as an indirect steering mechanism. We call this approach \textbf{R}einforcement \textbf{L}earning for \textbf{S}ource \textbf{R}ewriting (\textbf{RLSR}; see Figure~\ref{fig:overview}).
Because source rewriting aims to improve translation quality, we define the reward as the improvement in downstream translation quality: the difference in automatic metric scores between translations generated from the rewritten and original sources. We then train the rewriting model with an on-policy reinforcement learning (RL) algorithm.
Because RLSR optimizes a separate rewriting model for each MT model, serving many MT models would require many specialists; we therefore further show that these per-MT models can be combined, through a simple model-merging method, into a single universal rewriting model that serves all MT models at once.

We evaluate RLSR against a range of baselines on six MT models across all 16 language pairs in the WMT2025 General MT Shared Task \cite{wmt25}.
The results show that our 4B RLSR-trained rewriting models yield significant gains over both the no-rewriting baseline and prompt-based rewriting with the same 4B backbone, and are competitive with prompt-based rewriting driven by a 235B LLM.
\rev{A single rewriting model obtained by merging the six per-MT specialists preserves these conclusions.
Moreover, a specialist trained for one MT model can also improve a different MT model, and merging only specialists trained for other MT models can still improve a target whose own specialist is excluded.
These findings provide preliminary evidence that the learned rewriting behavior can transfer to unseen downstream MT models.}
Further analysis shows that, compared with rewriting models trained via supervised fine-tuning (SFT), RL-trained models generate more diverse rewrites and are less prone to degenerating into simple copies of the original source text.

Overall, our study shows that source rewriting can be made substantially more reliable by directly optimizing rewrites for downstream translation quality, offering new insight into how external rewriting models can improve black-box MT systems without modifying them.

\section{Related Work}
\label{sec:related_work}

Source rewriting is a well-studied area of MT research, with various strategies evaluated since the statistical MT era \cite{ape1,ape2,ape3}. 
These strategies include word reordering \cite{reorder1,reorder2,reorder3}, paraphrasing \cite{ape_pp1,ape_pp2,ape_pp3,rewriting}, simplification \cite{ape_simpo1,ape_simpo2,ape_bt,rewriting}, and conversion to translationese \cite{ape_bt}. 
However, some studies have questioned the practical effectiveness of certain rewriting strategies \cite{ape_eval1,ape_eval2}. 
Other work has analyzed which aspects of MT outputs are affected by source rewriting \cite{fujita21,fujita24}, as well as the conditions under which rewriting is appropriate and effective \cite{fujita17}.

Given the strong text-generation capabilities of LLMs \cite{gpt3,gpt4}, recent work has explored their potential for source rewriting. 
\citet{rewriting} investigated the effectiveness of three prompt strategies. 
In a complementary direction, \citet{rewriting_qe} proposed an approach that uses feedback from a segment-level, reference-free evaluation model at inference time, enabling the LLM to iteratively rewrite the source text. 
However, these methods share a key limitation: they control the rewriting model only \emph{indirectly}, through hand-crafted prompts (and, for \citet{rewriting_qe}, prompt-guided iterative refinement), without updating the model's parameters to optimize downstream translation quality.
Their effectiveness therefore hinges on how well an off-the-shelf LLM follows the prompt, a fragile dependence that, as we show, can break down for smaller models.
The inference-time approach of \citet{rewriting_qe} also incurs a high inference cost, since both the rewriting and MT models must be run repeatedly.
In contrast, RLSR optimizes the rewriting model's parameters directly for downstream translation quality and remains efficient at inference time, requiring only a single pass through the rewriting model and the downstream MT model.

\rev{\section{Problem Formulation}}
\label{sec:sr}
Source rewriting aims to improve the quality of the translations produced by a given MT model: it generates an alternative, meaning-preserving source text that is more likely to yield a high-quality translation, while leaving the MT model itself unchanged. 
Formally, let $s$ denote a source text and $r$ denote its reference translation. 
Let $M$ be a fixed MT model that maps source texts to translations, and let $R_{\theta}$ be a source rewriting model parameterized by $\theta$. 
Given $s$, the rewriting model generates a rewritten source text $\tilde{s} \sim R_{\theta}(\cdot \mid s)$.
The MT model then produces translations for both the original source, $M(s)$, and the rewritten source, $M(\tilde{s})$.

Let $Q(s, t, r)$ be an automatic evaluation function that scores a candidate translation $t$ against the original source $s$ and its reference translation $r$, with higher values indicating better translation quality. 
To quantify the utility of a rewrite, we measure the marginal improvement it yields in the final translation. 
We define this improvement as the reward $\mathcal{R}$:
\begin{equation}
     \mathcal{R}(s, \tilde{s}, r) = Q(s, M(\tilde{s}), r) - Q(s, M(s), r).
     \label{eq:reward}
\end{equation}
The goal of source rewriting is to learn a rewriting model $R_{\theta}$ that yields rewrites with positive reward ($\mathcal{R} > 0$).
When the rewriting model is an LLM without task-specific rewriting training, its behavior is controlled only indirectly, through natural-language prompts intended to elicit high-reward rewrites \cite{rewriting,rewriting_qe}; the reward $\mathcal{R}$ is never observed during generation, so the model receives no feedback on whether a given rewrite actually helps.
Consequently, as shown in Section~\ref{sec:exp_results}, prompt-based rewriting can even yield negative reward ($\mathcal{R} < 0$), especially for smaller LLMs, whose limited instruction-following ability makes them more prone to producing harmful rewrites.

\section{Proposed Methods}
\label{sec:proposed_methods}
\subsection{RLSR}
\label{sec:rlsr}
To overcome this limitation, we optimize the rewriting model directly to maximize the reward $\mathcal{R}(s, \tilde{s}, r)$ defined in Eq.~\ref{eq:reward}.
However, this objective cannot be maximized by standard SFT: the downstream MT model $M$, the evaluation metric $Q$, and the discrete generation process for $\tilde{s}$ are all non-differentiable, so gradients cannot be backpropagated from the reward to the rewriting model.
A natural solution is to formulate this optimization as an RL problem.
We therefore propose \textbf{RLSR}, which uses improvements in downstream translation quality directly as rewards.

RLSR starts from an initial rewriting model, which we denote as the reference policy $R_{\text{ref}}$ (e.g., an instruction-following LLM given a rewriting prompt). 
During training, given a source text $s$ sampled from a dataset $\mathcal{D} = \{(s, r)\}$, the current rewriting model $R_{\theta}$ acts as the policy and samples a rewrite $\tilde{s} \sim R_{\theta}(\cdot|s)$. 
The fixed downstream MT model $M$ then translates both $s$ and $\tilde{s}$. 
The two translations are then evaluated with the metric $Q$ to compute the reward $\mathcal{R}(s, \tilde{s}, r)$. 

To discourage reward hacking (e.g., generating nonsensical text that exploits metric vulnerabilities), we anchor the model to the reference policy $R_{\text{ref}}$ using a Kullback--Leibler (KL) penalty. 
The adjusted total reward $\mathcal{R}_{\text{total}}$ is defined as:
\begin{equation}
    \mathcal{R}_{\text{total}}(s, \tilde{s}, r) = \mathcal{R}(s, \tilde{s}, r) - \beta \log \frac{R_{\theta}(\tilde{s}|s)}{R_{\text{ref}}(\tilde{s}|s)},
    \label{eq:total_reward}
\end{equation}
where $\beta$ is a coefficient controlling the strength of the KL penalty,
and $R_{\theta}(\tilde{s}|s)$ denotes the sequence-level probability assigned by the model to the rewrite $\tilde{s}$ conditioned on $s$. 

The RLSR objective is to maximize the expected adjusted reward over the training dataset:
\begin{equation}
    J(\theta) = \mathbb{E}_{(s, r) \sim \mathcal{D}, \, \tilde{s} \sim R_{\theta}(\cdot|s)} \left[ \mathcal{R}_{\text{total}}(s, \tilde{s}, r) \right]. 
\end{equation}
To optimize $J(\theta)$, we employ Decoupled Clip and Dynamic Sampling Policy Optimization (DAPO) \cite{dapo}, an on-policy RL algorithm. 
Implementation details are provided in Appendix~\ref{sec:rlsr_details}.

\rev{\subsection{A Universal Rewriting Model}
\label{sec:ensemble}
RLSR (Section~\ref{sec:rlsr}) optimizes a rewriting model for a single downstream MT model $M$.
In many practical settings, however, we would like a single rewriting model that improves translation across several MT models at once.
A direct way to pursue this goal is to optimize the rewriting model for the aggregated quality improvement over a set of $K$ MT models $\{M_1, \dots, M_K\}$, i.e., to maximize the joint reward:
\begin{align}
    \mathcal{R}_{\text{joint}}(s, \tilde{s}, r) &=
    \sum_{k=1}^{K} \big[\, Q\big(s, M_k(\tilde{s}), r\big) \nonumber
    \\ & - Q\big(s, M_k(s), r\big)\,\big],
    \label{eq:joint_reward}
\end{align}
and to train one model against this objective with RLSR; we refer to this strategy as \emph{joint training}.

Although conceptually simple, joint training scales poorly with $K$: each reward computation requires decoding with all $K$ MT models and scoring each output with the metric, which becomes prohibitive as $K$ grows.
We therefore pursue the same objective indirectly, by merging the per-MT rewriting models that RLSR has already produced into a single model.
Suppose we have $K$ specialist rewriting models, each trained with RLSR for one MT model and all fine-tuned from the same base model $R_{\text{ref}}$.
We combine them into a single universal rewriting model by merging their parameters and weighting all specialists equally.
The resulting model requires no further training or access to the MT models; we call this merged model \textbf{RLSR-Ens}.
To perform the merge, we use DARE~\cite{dare}, which sparsifies and rescales each specialist's task vector before averaging the task vectors and adding the result back to the base model; we defer the full mathematical definition to Appendix~\ref{sec:dare_details}.}

\section{Experiments}

\subsection{Experimental Setup}
\label{sec:exp_setup}

\paragraph{Training Data}
We constructed the RLSR training data from the WMT2019--2024 News/General MT Shared Task test sets \cite{wmt19,wmt20,wmt21,wmt22,wmt23,wmt24}.\footnote{As of WMT2022, the news MT task was renamed to the general MT task.} 
The resulting dataset comprises 21K samples across 24 language pairs. 
We randomly selected 5\% of the data for validation and used the remainder for training.

\paragraph{Models}
Our downstream MT models ($M$) come from two categories of LLMs. The first category consists of general-purpose LLMs: Gemma3 27B \cite{gemma3},
Gemma4 31B,\footnote{\url{https://huggingface.co/google/gemma-4-31B-it}}   
Qwen3 30B \cite{qwen3},
and Qwen3 4B \cite{qwen3}.
The second category consists of translation-optimized models: Translategemma 27B  \cite{translategemma}
and an in-house Qwen3 4B SFT-RL model.\footnote{This MT model was developed by us using a two-stage training process comprising SFT and DAPO. Training details are provided in Appendix~\ref{sec:qwen_sft_rl}.} 
For the source rewriting models, we used Qwen3 4B and Qwen3 235B \cite{qwen3}. We focused on the Qwen3 family because a preliminary study with the Gemma4 family showed that prompt-based rewriting did not yield consistent improvements (Appendix~\ref{sec:gemma4}).
Because of computational constraints, we trained RLSR only with Qwen3 4B. 
The prompts used for the MT and rewriting models are given in Appendix~\ref{sec:prompt}.

\paragraph{Training and Inference Configuration}
We used \texttt{xCOMET-XL} \cite{xcomet} as the training metric $Q$ because it correlates strongly with human evaluation \cite{wmtm23,wmtm24} while remaining computationally feasible; further training details are provided in Appendix~\ref{sec:rlsr_details}.
\rev{For RLSR-Ens, we merged the six per-MT RLSR specialists, setting the task-vector density to 0.5.}
At inference time, we used greedy decoding for all MT and rewriting models.

\paragraph{Baselines}
We compared RLSR against the following baselines:\footnote{We did not include the method of \citet{rewriting_qe} as a baseline because their code is not publicly available, and the paper does not provide sufficient detail to reproduce it; in Appendix~\ref{sec:qe_selective}, we instead combined RLSR with the core idea behind their approach.} 
\begin{itemize}
    \item \textbf{No-rewriting}: The original source text was fed directly to the MT models for translation without modification.
    \item \textbf{Prompt-based Rewriting}: We used three prompt strategies proposed by \citet{rewriting}: \textbf{Simplification}, \textbf{Paraphrase}, and \textbf{Easy Translate}. These prompts instruct LLMs (without task-specific rewriting training) to simplify the text, paraphrase it, or rewrite it into a more translatable form. We adopted these prompts because, to the best of our knowledge, they are the only prompts proposed in prior work specifically for LLM-based source rewriting. We instantiated these prompt-based baselines with LLMs at two scales: Qwen3 4B and Qwen3 235B. The prompts are detailed in Appendix~\ref{sec:prompt}.
\end{itemize}

\paragraph{Evaluation}
We conducted our evaluation on the complete test set of the WMT2025 General MT Shared Task \cite{wmt25}, which covers 16 language pairs. 
For evaluation, we used two learned reference-based metrics, xCOMET (\texttt{xCOMET-XXL}) \cite{xcomet} 
and MetricX (\texttt{MetricX-24-XXL}) \cite{metricx24}, 
as well as GEMBA-MQM \cite{gemba-mqm} (henceforth GEMBA), a prompt-based, reference-free LLM metric based on \texttt{Gemini 3.0 Flash} \cite{gemini}.
We selected these metrics because they correlate strongly with human evaluation \cite{wmtm23,wmtm24}.
However, because a learned metric (\texttt{xCOMET-XL}) was used explicitly as the reward for optimizing the rewriting model during RLSR training, there is a risk of ``metric bias'' \cite{metric_bias}: other learned metrics may also favor translations generated from RLSR-rewritten sources. 
To obtain a more independent check, we also evaluated our method with GEMBA.
We performed paired bootstrap tests \cite{pbs} to compare \rev{both RLSR and RLSR-Ens} with the prompt-based rewriting baselines that use the same base model (Qwen3 4B), as well as with the no-rewriting baseline.

\begin{table*}[t]
\centering
\small
\setlength\tabcolsep{3.5pt}
\begin{tabular}{llcccccc}
\toprule
\multirow{2}{*}{Rewriting Model} & \multirow{2}{*}{Rewriting Prompt} & \multicolumn{6}{c}{Downstream MT Model} \\
\cmidrule(lr){3-8}
 & & xCOMET$\uparrow$ & MetricX$\downarrow$ & GEMBA$\downarrow$ & xCOMET$\uparrow$ & MetricX$\downarrow$ & GEMBA$\downarrow$ \\
\midrule
& & \multicolumn{3}{c}{Gemma3 27B} & \multicolumn{3}{c}{Gemma4 31B} \\
\midrule
- & - & 49.22 & 7.19 & 8.73 & 51.73& 6.83&6.38 \\
\hdashline
Qwen3 4B & Simplification & 47.66 & 7.12 & 10.01& 49.07& 6.97& 6.91\\
Qwen3 4B & Paraphrase & 43.85 & 7.18 & 9.14 &44.85 &7.02 & 7.54\\
Qwen3 4B & Easy Translate & 48.85 & 7.10 & 8.99&50.45 & 6.75& 7.01\\
\hdashline
Qwen3 235B & Simplification &49.92 & 6.82&8.18 & 51.43& 6.63& 6.19\\
Qwen3 235B & Paraphrase & 44.43& 7.01& 9.82& 45.67& 6.87& 8.27\\
Qwen3 235B & Easy Translate &\textbf{50.08} & \textbf{6.63}& \textbf{7.23}& 51.41&\textbf{6.54} & 6.09\\
\hdashline
Qwen3 4B (RLSR) & Easy Translate & ~~50.04\textsuperscript{†}& ~~6.99\textsuperscript{†}& ~~7.36\textsuperscript{†}& ~~\textbf{52.35}\textsuperscript{†}& ~~6.64\textsuperscript{†}&~~\textbf{6.00}\textsuperscript{†} \\
\rev{Qwen3 4B (RLSR-Ens)} & Easy Translate & ~~49.98\textsuperscript{†}& ~~7.01\textsuperscript{†}& ~~7.31\textsuperscript{†}& ~~52.32\textsuperscript{†}& ~~6.69\textsuperscript{†}&~~6.08\textsuperscript{†} \\
\midrule
& & \multicolumn{3}{c}{Translategemma 27B} & \multicolumn{3}{c}{Qwen3 30B} \\
\midrule
- & - & 57.45 & 5.11 & 5.97 &41.25 & 9.62& 15.62\\
\hdashline
Qwen3 4B & Simplification & 52.62 & 5.79 & 6.23& 40.20& 9.26& 16.10\\
Qwen3 4B & Paraphrase & 48.15 & 5.83 & 6.99 &37.19 &9.33 & 17.08\\
Qwen3 4B & Easy Translate & 54.87 & 6.54 & 6.15&40.97 & 9.30& 15.54\\
\hdashline
Qwen3 235B & Simplification & 55.62& 5.38&6.33 &42.16 &\textbf{8.97} & 15.03\\
Qwen3 235B & Paraphrase &49.12 &5.57 & 7.00&38.00 &9.18 & 16.55\\
Qwen3 235B & Easy Translate &55.35 &5.33 & 6.01&\textbf{42.25} & 9.01& 13.98\\
\hdashline
Qwen3 4B (RLSR) & Easy Translate & ~~\textbf{57.90}\textsuperscript{†}& \textbf{5.04}&~~5.69\textsuperscript{†} &~~42.21\textsuperscript{†} &~~9.21\textsuperscript{†} &~~\textbf{13.92}\textsuperscript{†} \\
\rev{Qwen3 4B (RLSR-Ens)} & Easy Translate & ~~57.75\textsuperscript{†}& 5.09& ~~\textbf{5.62}\textsuperscript{†}& ~~41.98\textsuperscript{†}& ~~9.33\textsuperscript{†}&~~14.05\textsuperscript{†} \\
\midrule
& & \multicolumn{3}{c}{Qwen3 4B} & \multicolumn{3}{c}{Qwen3 4B SFT-RL} \\
\midrule
- & - & 32.88 & 12.37 & 22.81 & 36.12&10.95 & 19.05\\
\hdashline
Qwen3 4B & Simplification & 31.71  & 14.02 & 23.16& 34.70&10.80 &19.21 \\
Qwen3 4B & Paraphrase & 31.05 & 14.94 & 23.45& 32.93& 10.69& 19.86\\
Qwen3 4B & Easy Translate & 32.79 & 13.06 & 21.99 & 36.02& 10.72& 19.19\\
\hdashline
Qwen3 235B & Simplification & 33.43 & 11.95 & 20.76 & 36.46& 10.53& 18.94\\
Qwen3 235B & Paraphrase & 32.12 & 14.14 & 22.97 & 33.84&10.59 & 18.78\\
Qwen3 235B & Easy Translate & 33.77 & 11.41 & 20.10 & \textbf{36.85}& 10.31& \textbf{17.01}\\
\hdashline
Qwen3 4B (RLSR) & Easy Translate & ~~34.02\textsuperscript{†} & ~~\textbf{11.19}\textsuperscript{†} & ~~\textbf{19.94}\textsuperscript{†} & ~~36.40\textsuperscript{†}&~~\textbf{10.25}\textsuperscript{†} & ~~17.22\textsuperscript{†}\\
\rev{Qwen3 4B (RLSR-Ens)} & Easy Translate & ~~\textbf{34.09}\textsuperscript{†}& ~~11.24\textsuperscript{†}& ~~19.99\textsuperscript{†}& ~~36.45\textsuperscript{†}& ~~10.27\textsuperscript{†}&~~17.45\textsuperscript{†} \\
\bottomrule
\end{tabular}
\caption{Comparison of rewriting methods on the WMT2025 General MT Shared Task. We performed a paired bootstrap test comparing each of Qwen3 4B (RLSR) \rev{and Qwen3 4B (RLSR-Ens)} against both the Qwen3 4B prompt-based rewriting baselines and the no-rewriting baseline; † indicates significantly better performance than all those baselines ($p < 0.05$).}
\label{tab:main_exp1}
\end{table*}

\subsection{Experimental Results}
\label{sec:exp_results}

Table~\ref{tab:main_exp1} reports evaluation results on the WMT2025 General MT Shared Task.
Results grouped by the source language are provided in Appendix~\ref{sec:src_lang_group_exp}.

\paragraph{Prompt-based Rewriting Degrades Quality at 4B Scale}
For all six downstream MT models, every Qwen3 4B prompt-based baseline performs worse than the no-rewriting baseline on xCOMET. 
GEMBA shows a largely similar pattern---with the sole exceptions of the Easy Translate baseline for Qwen3 30B and Qwen3 4B---while MetricX is mixed but shows no consistent gain.
In short, prompting a 4B LLM to rewrite the source, does not reliably improve translation quality and frequently degrades it. This is precisely the failure mode we attribute to controlling rewriting through prompts alone.

\paragraph{RLSR vs. No-Rewriting and Same-Scale Baselines}
The paired bootstrap tests show that RLSR's gains over the Qwen3 4B prompt-based baselines and the no-rewriting baseline are significant in most settings.
One minor exception is that, when using Translategemma 27B, RLSR does not significantly outperform all relevant baselines on MetricX.
However, because the improvements on xCOMET and GEMBA for this MT model remain significant, this isolated exception does not change our overall conclusion.
Overall, RLSR significantly outperforms the no-rewriting baseline and same-scale prompt-based rewriting baselines.
\rev{The RLSR-Ens model shows the same overall pattern as the per-MT RLSR models.}

\paragraph{RLSR vs. 235B LLM Baselines}
Across all metrics, the RLSR-trained 4B models achieve performance comparable to that of prompt-based rewriting baselines using a 235B LLM (Qwen3 235B).
This result suggests that explicitly optimizing a compact rewriting model for downstream translation quality provides a parameter-efficient alternative to increasing model size.

\paragraph{Metric Bias}
As noted in Section~\ref{sec:exp_setup}, training directly against \texttt{xCOMET-XL} raises concerns about potential metric bias.
However, our RLSR model consistently achieves lower (better) GEMBA scores than the no-rewriting baseline across all six MT models. 
Because GEMBA is a prompt-based LLM metric that operates on principles fundamentally different from those of the learned xCOMET metric, these consistent improvements suggest that our model does not merely overfit the reward but instead improves the underlying translation quality.

\section{Discussion}

\subsection{Why RL? A Comparison with SFT}
\label{sec:why_rl}

\paragraph{Training Rewriting Models via SFT} 
Although SFT cannot directly maximize the non-differentiable reward, it can be applied to an offline dataset of $(s, \tilde{s})$ pairs selected to approximate reward maximization. For each original source $s$, we first use the initial model $R_{\text{ref}}$ to generate $N$ distinct rewriting candidates $\{\tilde{s}_1, \tilde{s}_2, \dots, \tilde{s}_N\}$. We then define the supervised target as the highest-reward candidate in a set that also includes the unmodified source: $\tilde{s}^* \in \operatorname*{arg\,max}_{\hat{s} \in \mathcal{C}(s)} \mathcal{R}(s, \hat{s}, r),$ where $\mathcal{C}(s) = \{s, \tilde{s}_1, \tilde{s}_2, \dots, \tilde{s}_N\}$.
The model is then trained on the pair $(s, \tilde{s}^*)$ using the standard negative log-likelihood SFT objective: $\mathcal{L}_{\text{SFT}} = -\sum_{t=1}^{|\tilde{s}^*|} \log R_{\theta}(\tilde{s}^*_t | s, \tilde{s}^*_{<t})$.

\paragraph{Experimental Setup} 
For a fair comparison, the SFT baselines used the same base model (Qwen3 4B), rewriting prompt (``Easy Translate''), reward (\texttt{xCOMET-XL}), and source-text distribution as RLSR. 
Because preliminary SFT runs performed poorly, we restricted our SFT experiments to two MT models: Gemma3 27B and Qwen3 30B. 
Further training details are provided in Appendix~\ref{sec:sft}.

\begin{table}[h]
\centering
\setlength\tabcolsep{2pt}
\small
\begin{tabular}{llcccccc}
\toprule
MT Model & Training & xCOMET & MetricX & GEMBA \\
\midrule
\multirow{4}{*}{Gemma3} & -  & 49.22 & 7.19 & 8.73 \\
& RLSR & \textbf{50.04}& \textbf{6.99}& \textbf{7.36}\\
& SFT & 48.41&7.89 & 9.05\\
& SFT (filtered) & 46.82&9.66 & 10.63\\
\midrule
\multirow{4}{*}{Qwen3} & - & 41.25 & 9.62& 15.62\\
& RLSR & \textbf{42.21} &\textbf{9.21} &\textbf{13.92}\\
& SFT & 41.19&9.66 & 15.88\\
& SFT (filtered) & 40.95&9.72 & 16.01\\
\bottomrule
\end{tabular}
\caption{Performance comparison between rewriting models trained via RLSR and SFT. ``Gemma3'' and ``Qwen3'' refer to Gemma3 27B and Qwen3 30B, respectively. ``SFT (filtered)'' denotes training on a subset where unchanged targets ($\tilde{s}^* = s$)  are removed.}
\label{tab:sft_exp}
\end{table}

\paragraph{Performance Degradation under SFT}
The evaluation results in Table~\ref{tab:sft_exp} show that the SFT-trained models consistently underperform their RL-trained counterparts across all metrics and also fall below the no-rewriting baseline. 

\paragraph{Copying Degeneration}
To understand this degradation, we measured how often each model copies the source verbatim (i.e., $\tilde{s}=s$).
SFT-trained models copy the source in 89.74\% and 75.66\% of cases for Gemma3 27B and Qwen3 30B, respectively, whereas the RL-trained models do so in only 6.33\% and 8.09\% of cases.
Yet unchanged targets constitute only 10.10\% and 13.98\% of the corresponding SFT training sets, so they do not explain this behavior.
A more plausible explanation lies in the surface form of the supervised targets: most reward-improving rewrites change only a few words or split a long sentence, leaving most tokens unchanged; SFT's token-level cross-entropy thus makes copying an easy local optimum.
Copies instead receive zero reward under RL, encouraging more varied and targeted rewrites.

\paragraph{Filtering Unchanged Sources}
To test whether removing these unchanged pairs would force the SFT model to learn meaningful strategies, we trained another model on a filtered dataset (``SFT (filtered)'' in Table~\ref{tab:sft_exp}) that excludes all $\tilde{s}=s$ instances. 
However, performance deteriorated further. 
We attribute this to the equal weighting of all tokens in the SFT objective. As noted above, reward-improving rewrites alter only a small fraction of the source, so the few tokens that drive the improvement are vastly outnumbered by tokens carried over unchanged. The equally weighted SFT loss cannot isolate these crucial tokens within highly similar sequences. 
RL, by contrast, trains the model to differentiate trajectories with nearly identical tokens but substantially different rewards, allowing it to focus on the crucial modifications.

In summary, RL provides a more effective paradigm for training rewriting models than SFT. 
As a middle ground, we also explored Direct Preference Optimization (DPO), which, like RL, learns to discriminate between similar rewrites of differing quality. DPO improves over SFT but remains markedly less stable than RLSR (Appendix~\ref{sec:dpo}).
Advanced SFT techniques, such as identifying critical tokens and assigning them larger weights in the training objective \cite{token_sft}, might narrow this gap; we leave this direction to future work.

\rev{\subsection{Rewriting Across Multiple MT Models}
\label{sec:cross_mt}
We study both cross-MT generalization and approaches to serving multiple MT models, using Gemma3 27B and Qwen3 30B as test cases.

\paragraph{Cross-Application}
We apply each specialist trained for one target to the other without further training.
As Table~\ref{tab:cross_mt_exp} shows, both cross-applied specialists improve over no rewriting on all metrics and remain close to the target-specific specialists.

\paragraph{Leave-One-MT-Out Merging}
We also evaluate \textbf{RLSR-Ens-LOO}, which merges five specialists while excluding the one trained for the target MT model.
RLSR-Ens-LOO improves over no rewriting on all three metrics for both Gemma3 and Qwen3.
Together, these results provide preliminary evidence of generalization to unseen downstream MT models.}

\begin{table}[t]
\centering
\setlength\tabcolsep{2pt}
\small
\begin{tabular}{llccc}
\toprule
MT Model & \footnotesize{Rewriting Model} & xCOMET & MetricX & GEMBA \\
\midrule
\multirow{6}{*}{Gemma3} & -  & 49.22 & 7.19 & 8.73 \\
& For Gemma3 & 50.04& 6.99& 7.36\\
& For Qwen3 & \textbf{50.16}&6.95 & 7.11\\
& For Both & 50.10&\textbf{6.84} & \textbf{7.08}\\
& \rev{RLSR-Ens} & 49.98 & 7.01 & 7.31\\
& \rev{RLSR-Ens-LOO} & 50.14 & 6.99 & 7.37\\
\midrule
\multirow{6}{*}{Qwen3} & - & 41.25 & 9.62& 15.62\\
& For Qwen3 & 42.21 &9.21 &13.92\\
& For Gemma3 & 42.07&9.24 & 14.07\\
& For Both & \textbf{42.25}&\textbf{9.17} & \textbf{13.88}\\
& \rev{RLSR-Ens} & 41.98 & 9.33 & 14.05\\
& \rev{RLSR-Ens-LOO} & 41.86 & 9.33 & 14.01\\
\bottomrule
\end{tabular}
\caption{Rewriting across multiple MT models. ``For X'' denotes a specialist trained for X, ``For Both'' denotes joint training, RLSR-Ens merges all six specialists, and RLSR-Ens-LOO merges the five non-target specialists.}
\label{tab:cross_mt_exp}
\end{table}

\rev{\paragraph{Joint Training vs.\ Merging}
We finally compare the two routes to a single model that serves multiple represented MT models.
Joint training (``For Both''), which directly optimizes the joint reward of Eq.~\ref{eq:joint_reward}, improves both MT models simultaneously and performs comparably to the per-MT specialists.
The DARE merge (``RLSR-Ens'') is nearly as strong, recovering most of the per-MT gain on both MT models.
The two routes differ sharply in scalability: joint training must run all $K$ MT models and the metric at every step, so its cost grows with $K$, and our compute budget therefore limited it to $K=2$.
Merging adds no per-step MT or metric cost and is therefore a low-cost, scalable alternative to joint training for building a universal rewriting model.}

\subsection{What Does RLSR Learn to Rewrite?}
\label{sec:case_study}
\begin{table}[h]
\centering
\setlength\tabcolsep{2pt}
\small
\begin{tabular}{lccc}
\toprule
Method & Unchanged & Len.\ ratio & Src.\ sim. \\
\midrule
Simplification (235B)  & 0.00\% & 0.90 & 0.19 \\
Paraphrase (235B)      & 0.00\% & 1.13 & 0.09 \\
Easy Translate (235B)  & 0.05\% & 0.98 & 0.17 \\
\hdashline
RLSR (4B, Gemma3)      & 6.33\% & 1.00 & 0.85 \\
RLSR (4B, Qwen3)       & 8.09\% & 1.00 & 0.69 \\
\rev{RLSR-Ens (4B)}       & 9.47\% & 1.00 & 0.90 \\
\bottomrule
\end{tabular}
\caption{Extent of source alteration by each method. ``Unchanged'': fraction of sources copied verbatim; ``Len.\ ratio'': median character length of the rewrite divided by that of the source; ``Src.\ sim.'': median character-level similarity (\texttt{difflib}) between rewrite and source.}
\label{tab:edit_locality}
\end{table}
Beyond aggregate scores, we qualitatively examine \emph{how} RLSR rewrites differ from prompt-based rewrites. 
Throughout this section, the prompt-based baselines refer to the strongest variants in Table~\ref{tab:main_exp1}, i.e., those using the 235B LLM, whereas RLSR uses its 4B model.
In the case-study examples, the downstream MT model is Qwen3~30B.

\paragraph{RLSR Makes Localized, Length-Preserving Edits}
We first quantify how much each method alters the source.
Table~\ref{tab:edit_locality} reports the fraction of sources left exactly unchanged, the median rewrite/source length ratio, and the median character-level similarity to the source (see Appendix~\ref{sec:src_sim_details} for details). 
The prompt-based methods rewrite the input almost wholesale, with source similarities of only $0.09$--$0.19$. 
Paraphrase expands the text (length ratio $1.13$), whereas Simplification compresses it ($0.90$).
In contrast, RLSR keeps the length essentially unchanged (ratio $1.00$) and retains most of the source ($0.69$--$0.85$ similarity), editing it only locally; the merged RLSR-Ens behaves the same way and is, if anything, the most conservative of all ($9.47\%$ unchanged, $0.90$ similarity).
RLSR therefore avoids two failure modes: the ``copying'' degeneration of SFT (Section~\ref{sec:why_rl}), which leaves $75$--$90\%$ of sources unchanged, and the indiscriminate full rewriting of prompt-based methods.

\begin{table*}[t]
\centering
\small
\renewcommand{\arraystretch}{1.15}
\begin{tabularx}{\textwidth}{@{}l X@{}}
\toprule
\multicolumn{2}{@{}l}{\textbf{Example 1 (en$\rightarrow$zh).} The source is already fluent; only source-side rewrites are shown.} \\
\midrule
Source (no-rewriting)           & Like some kind of \textcolor{red}{infantilised diaper king}, Trump has genuinely had a special Diet Coke-summoning button installed in the Oval Office \\
Simplification    & Like a \textcolor{red}{childish diaper-wearing king}, Trump really had a special button in the Oval Office to call for Diet Coke. \emph{(loaded diction flattened)} \\
Paraphrase        & Trump has actually installed a dedicated button in the Oval Office designed specifically to call for his favorite Diet Coke, \textcolor{red}{a whimsical touch that evokes the indulgence of a pampered figure, almost like a grown-up version of a child's fantasy.} \emph{(fabricated)} \\
Easy Translate    & Trump really had a special button installed in the Oval Office to call for Diet Coke, like a \textcolor{red}{childish king who wears diapers}. \emph{(loaded diction flattened)}\\
\textbf{RLSR}     & Like some kind of \textcolor{red}{infantilized diaper king}, Trump has genuinely had a special Diet Coke-summoning button installed in the Oval Office. \emph{(only the spelling is changed; RLSR-Ens makes the identical edit)} \\
\midrule
\multicolumn{2}{@{}l}{\textbf{Example 2 (en$\rightarrow$zh).} A disfluent, spoken-style source (excerpted).} \\
\midrule
Source (no-rewriting)           & \ldots\ only covers \textcolor{red}{x amount 30 days or whatever} \ldots\ the \textcolor{red}{area} store \ldots\ the \textcolor{red}{area} hotline \ldots \\
\textbf{RLSR}     & \ldots\ covers \textcolor{red}{a certain amount, \ldots\ 30 days or something} \ldots\ the \textcolor{red}{local} store \ldots\ the \textcolor{red}{local} hotline \ldots \\
\textbf{RLSR-Ens} & \ldots\ covers \textcolor{red}{a certain amount, \ldots\ 30 days or so} \ldots\ the \textcolor{red}{area} store \ldots\ the \textcolor{red}{area} hotline \ldots \\
\hdashline
MT (no-rewriting) & \zh{\ldots 30天或\textcolor{red}{whatever}\ldots \textcolor{red}{当地}门店\ldots \textcolor{red}{区域}热线\ldots} \emph{(``whatever'' untranslated; ``area'' inconsistent)} \\
MT (\textbf{RLSR})& \zh{\ldots 30天左右\ldots \textcolor{red}{当地}门店\ldots \textcolor{red}{当地}热线\ldots} \emph{(fluent and consistent)} \\
MT (\textbf{RLSR-Ens})& \zh{\ldots 30天左右\ldots \textcolor{red}{当地}门店\ldots \textcolor{red}{区域}热线\ldots} \emph{(``whatever'' fixed; ``area'' inconsistent, as in no-rewriting)} \\
\bottomrule
\end{tabularx}
\caption{Case study comparing RLSR and the merged RLSR-Ens with the prompt-based methods. \textcolor{red}{Red} spans mark the key differences.}
\label{tab:case_study}
\end{table*}

\paragraph{Case Study 1: RLSR Refrains from Rewriting an Already Adequate Source}
Example~1 in Table~\ref{tab:case_study} contrasts the four methods on a source sentence that is already fluent and readily translatable. 
Simplification and Easy Translate remain faithful to the source but flatten the author's deliberately loaded diction, replacing \textit{``infantilised diaper king''} with the blander \textit{``childish \ldots\ king''} and unpacking the compound \textit{``Diet Coke-summoning button''}. 
The Paraphrase baseline drifts even further: it drops the original \textit{``diaper king''} image and fabricates content that is absent from the source, namely \textit{``a whimsical touch that \ldots''}. The MT model then translates this unsupported elaboration as if it were genuine. 
This tendency to rewrite indiscriminately helps explain why, despite using a 235B LLM, Paraphrase falls below the no-rewriting baseline on almost all MT models in Table~\ref{tab:main_exp1}.
RLSR \rev{and RLSR-Ens instead} leave the sentence essentially intact, editing only the British spelling \textit{infantilised}~$\rightarrow$~\textit{infantilized} and preserving the meaning.

\paragraph{Case Study 2: RLSR Targets the Actual Translation Obstacles}
Example~2 in Table~\ref{tab:case_study} shows a disfluent, spoken-style source whose real obstacle to translation is the broken phrase \textit{``x amount 30 days or whatever''}: left unedited, Qwen3~30B copies the stray token \textit{``whatever''} verbatim into the Chinese output.
\rev{Both RLSR and RLSR-Ens normalize this phrase, so the untranslated token disappears.
The two models differ most notably on the ordinary word \textit{``area''}, which RLSR happens to rewrite to \textit{``local''} while the more conservative RLSR-Ens leaves it untouched.
This difference is immaterial: unlike the disfluent \textit{``whatever''}, \textit{``area''} is perfectly translatable, and Qwen3~30B renders it inconsistently (\zh{当地}, ``local'', vs.\ \zh{区域}, ``regional'') whether or not it is rewritten; the same inconsistency already appears in the no-rewriting output.
It is therefore a quirk of this particular MT model, not a deficiency of the rewrite, and is consistent with the preliminary cross-MT transfer observed in Section~\ref{sec:cross_mt}.}

\rev{\section{Conclusions}}
We introduced RLSR, a framework that trains a source rewriting model using each rewrite's improvement in downstream translation quality as the reward. This replaces the indirect control of natural-language prompts with direct optimization of the task objective.
Across six MT models and 16 language pairs, 4B RLSR models significantly outperform both the no-rewriting baseline and same-scale prompt-based rewriting methods, and match the performance of prompt-based baselines built on a 235B LLM.
\rev{We further show that a single universal rewriting model, obtained by merging the per-MT specialists, nearly matches the specialists across all six MT models, providing a low-cost way to serve them all at once.}
\rev{Applying a specialist to another MT model and merging only specialists trained for other MT models also suggest that the learned rewriting behavior may generalize to unseen downstream MT models.}


\section*{Limitations}
First, we assume that the automatic metrics we use faithfully reflect improvements in translation quality.
Because the RLSR reward is computed directly from \texttt{xCOMET-XL}, learned metrics such as xCOMET and MetricX may be biased in favor of RLSR-rewritten sources; as discussed in Section~\ref{sec:exp_results}, the consistent gains under the independent GEMBA metric mitigate this concern.
A more definitive answer would require human evaluation, which we did not conduct because of its substantial cost.
 
Second, our experiments are limited in scope. 
Although we cover six MT models and 16 language pairs, evaluating RLSR on a broader range of MT models, domains, and language pairs would further strengthen our conclusions.
 
\rev{Finally, RLSR is costly to train and to scale across MT models. Training is considerably more expensive than standard SFT, since it requires repeatedly running the downstream MT model and the evaluation metric to compute rewards. 
As a result, RLSR may be impractical when compute budgets are tight.}


\bibliography{custom}

@inproceedings{hpe1,
    title = "Pre-editing and the use of simplified writing for {MT}",
    author = "Pym, Peter",
    editor = "Mayorcas, Pamela",
    booktitle = "Proceedings of Translating and the Computer 10: The translation environment 10 years on",
    month = nov # " 10-11",
    year = "1988",
    address = "London, UK",
    publisher = "Aslib",
    url = "https://aclanthology.org/1988.tc-1.11/"
}

@inproceedings{hpe2,
    title = "Two in one {--} can it work? Readability and translatability by means of controlled language",
    author = "Reuther, Ursula",
    booktitle = "EAMT Workshop: Improving MT through other language technology tools: resources and tools for building MT",
    month = apr # " 13",
    year = "2003",
    address = "Budapest, Hungary",
    publisher = "European Association for Machine Translation",
    url = "https://aclanthology.org/2003.eamt-1.14/"
}

@inproceedings{hpe3,
    title = "A Large-Scale Evaluation of Pre-editing Strategies for Improving User-Generated Content Translation",
    author = "Seretan, Violeta  and
      Bouillon, Pierrette  and
      Gerlach, Johanna",
    editor = "Calzolari, Nicoletta  and
      Choukri, Khalid  and
      Declerck, Thierry  and
      Loftsson, Hrafn  and
      Maegaard, Bente  and
      Mariani, Joseph  and
      Moreno, Asuncion  and
      Odijk, Jan  and
      Piperidis, Stelios",
    booktitle = "Proceedings of the Ninth International Conference on Language Resources and Evaluation ({LREC}'14)",
    month = may,
    year = "2014",
    address = "Reykjavik, Iceland",
    publisher = "European Language Resources Association (ELRA)",
    url = "https://aclanthology.org/L14-1532/",
    pages = "1793--1799"
}

@inproceedings{fujita21,
    title = "Understanding Pre-Editing for Black-Box Neural Machine Translation",
    author = "Miyata, Rei  and
      Fujita, Atsushi",
    editor = "Merlo, Paola  and
      Tiedemann, Jorg  and
      Tsarfaty, Reut",
    booktitle = "Proceedings of the 16th Conference of the European Chapter of the Association for Computational Linguistics: Main Volume",
    month = apr,
    year = "2021",
    address = "Online",
    publisher = "Association for Computational Linguistics",
    url = "https://aclanthology.org/2021.eacl-main.132/",
    doi = "10.18653/v1/2021.eacl-main.132",
    pages = "1539--1550"
}

@inproceedings{rewriting,
    title = "Automatic Input Rewriting Improves Translation with Large Language Models",
    author = "Ki, Dayeon  and
      Carpuat, Marine",
    editor = "Chiruzzo, Luis  and
      Ritter, Alan  and
      Wang, Lu",
    booktitle = "Proceedings of the 2025 Conference of the Nations of the Americas Chapter of the Association for Computational Linguistics: Human Language Technologies (Volume 1: Long Papers)",
    month = apr,
    year = "2025",
    address = "Albuquerque, New Mexico",
    publisher = "Association for Computational Linguistics",
    url = "https://aclanthology.org/2025.naacl-long.542/",
    doi = "10.18653/v1/2025.naacl-long.542",
    pages = "10829--10856",
    ISBN = "979-8-89176-189-6"
}

@inproceedings{ape_bt,
  title={Simplify-Then-Translate: Automatic Preprocessing for Black-Box Translation},
  author={Sneha Mehta and Bahareh Azarnoush and Boris Chen and Avneesh Singh Saluja and Vinith Misra and Ballav Bihani and Ritwik K. Kumar},
  booktitle={AAAI Conference on Artificial Intelligence},
  year={2020},
  url={https://api.semanticscholar.org/CorpusID:211018622}
}

@inproceedings{fujita24,
    title = "Automatic Decomposition of Text Editing Examples into Primitive Edit Operations: Toward Analytic Evaluation of Editing Systems",
    author = "Yamaguchi, Daichi  and
      Miyata, Rei  and
      Fujita, Atsushi  and
      Kajiwara, Tomoyuki  and
      Sato, Satoshi",
    editor = "Calzolari, Nicoletta  and
      Kan, Min-Yen  and
      Hoste, Veronique  and
      Lenci, Alessandro  and
      Sakti, Sakriani  and
      Xue, Nianwen",
    booktitle = "Proceedings of the 2024 Joint International Conference on Computational Linguistics, Language Resources and Evaluation (LREC-COLING 2024)",
    month = may,
    year = "2024",
    address = "Torino, Italia",
    publisher = "ELRA and ICCL",
    url = "https://aclanthology.org/2024.lrec-main.170/",
    pages = "1899--1914"
}

@inproceedings{ape_pp1,
    title = "Improved Statistical Machine Translation Using Paraphrases",
    author = "Callison-Burch, Chris  and
      Koehn, Philipp  and
      Osborne, Miles",
    editor = "Moore, Robert C.  and
      Bilmes, Jeff  and
      Chu-Carroll, Jennifer  and
      Sanderson, Mark",
    booktitle = "Proceedings of the Human Language Technology Conference of the {NAACL}, Main Conference",
    month = jun,
    year = "2006",
    address = "New York City, USA",
    publisher = "Association for Computational Linguistics",
    url = "https://aclanthology.org/N06-1003/",
    pages = "17--24"
}

@inproceedings{ape_pp2,
    title = "Source-Language Entailment Modeling for Translating Unknown Terms",
    author = "Mirkin, Shachar  and
      Specia, Lucia  and
      Cancedda, Nicola  and
      Dagan, Ido  and
      Dymetman, Marc  and
      Szpektor, Idan",
    editor = "Su, Keh-Yih  and
      Su, Jian  and
      Wiebe, Janyce  and
      Li, Haizhou",
    booktitle = "Proceedings of the Joint Conference of the 47th Annual Meeting of the {ACL} and the 4th International Joint Conference on Natural Language Processing of the {AFNLP}",
    month = aug,
    year = "2009",
    address = "Suntec, Singapore",
    publisher = "Association for Computational Linguistics",
    url = "https://aclanthology.org/P09-1089/",
    pages = "791--799"
}

@inproceedings{ape_pp3,
    title = "Improved Statistical Machine Translation Using Monolingually-Derived Paraphrases",
    author = "Marton, Yuval  and
      Callison-Burch, Chris  and
      Resnik, Philip",
    editor = "Koehn, Philipp  and
      Mihalcea, Rada",
    booktitle = "Proceedings of the 2009 Conference on Empirical Methods in Natural Language Processing",
    month = aug,
    year = "2009",
    address = "Singapore",
    publisher = "Association for Computational Linguistics",
    url = "https://aclanthology.org/D09-1040/",
    pages = "381--390"
}

@inproceedings{ape_simpo1,
    title = "Can Text Simplification Help Machine Translation?",
    author = "{\v{S}}tajner, Sanja  and
      Popovic, Maja",
    booktitle = "Proceedings of the 19th Annual Conference of the {E}uropean Association for Machine Translation",
    year = "2016",
    url = "https://aclanthology.org/W16-3411/",
    pages = "230--242"
}

@inproceedings{ape_simpo2,
    title = "Automated Text Simplification as a Preprocessing Step for Machine Translation into an Under-resourced Language",
    author = "{\v{S}}tajner, Sanja  and
      Popovi{\'c}, Maja",
    editor = "Mitkov, Ruslan  and
      Angelova, Galia",
    booktitle = "Proceedings of the International Conference on Recent Advances in Natural Language Processing (RANLP 2019)",
    month = sep,
    year = "2019",
    address = "Varna, Bulgaria",
    publisher = "INCOMA Ltd.",
    url = "https://aclanthology.org/R19-1131/",
    doi = "10.26615/978-954-452-056-4_131",
    pages = "1141--1150"
}

@inproceedings{rewriting_qe,
    title = "Targeted Source Text Editing for Machine Translation: Exploiting Quality Estimators and Large Language Models",
    author = "Koretaka, Hyuga  and
      Fujita, Atsushi  and
      Kajiwara, Tomoyuki",
    editor = "Haddow, Barry  and
      Kocmi, Tom  and
      Koehn, Philipp  and
      Monz, Christof",
    booktitle = "Proceedings of the Tenth Conference on Machine Translation",
    month = nov,
    year = "2025",
    address = "Suzhou, China",
    publisher = "Association for Computational Linguistics",
    url = "https://aclanthology.org/2025.wmt-1.12/",
    doi = "10.18653/v1/2025.wmt-1.12",
    pages = "200--219",
    ISBN = "979-8-89176-341-8"
}

@inproceedings{wmt25,
    title = "Findings of the {WMT}25 General Machine Translation Shared Task: Time to Stop Evaluating on Easy Test Sets",
    author = "Kocmi, Tom  and
      Artemova, Ekaterina  and
      Avramidis, Eleftherios  and
      Bawden, Rachel  and
      Bojar, Ond{\v{r}}ej  and
      Dranch, Konstantin  and
      Dvorkovich, Anton  and
      Dukanov, Sergey  and
      Fishel, Mark  and
      Freitag, Markus  and
      Gowda, Thamme  and
      Grundkiewicz, Roman  and
      Haddow, Barry  and
      Karpinska, Marzena  and
      Koehn, Philipp  and
      Lakougna, Howard  and
      Lundin, Jessica  and
      Monz, Christof  and
      Murray, Kenton  and
      Nagata, Masaaki  and
      Perrella, Stefano  and
      Proietti, Lorenzo  and
      Popel, Martin  and
      Popovi{\'c}, Maja  and
      Riley, Parker  and
      Shmatova, Mariya  and
      Steingr{\'i}msson, Steinth{\'o}r  and
      Yankovskaya, Lisa  and
      Zouhar, Vil{\'e}m",
    editor = "Haddow, Barry  and
      Kocmi, Tom  and
      Koehn, Philipp  and
      Monz, Christof",
    booktitle = "Proceedings of the Tenth Conference on Machine Translation",
    month = nov,
    year = "2025",
    address = "Suzhou, China",
    publisher = "Association for Computational Linguistics",
    url = "https://aclanthology.org/2025.wmt-1.22/",
    doi = "10.18653/v1/2025.wmt-1.22",
    pages = "355--413",
    ISBN = "979-8-89176-341-8"
}

@inproceedings{
dapo,
title={{DAPO}: An Open-Source {LLM} Reinforcement Learning System at Scale},
author={Qiying Yu and Zheng Zhang and Ruofei Zhu and Yufeng Yuan and Xiaochen Zuo and YuYue and Weinan Dai and Tiantian Fan and Gaohong Liu and Juncai Liu and LingJun Liu and Xin Liu and Haibin Lin and Zhiqi Lin and Bole Ma and Guangming Sheng and Yuxuan Tong and Chi Zhang and Mofan Zhang and Ru Zhang and Wang Zhang and Hang Zhu and Jinhua Zhu and Jiaze Chen and Jiangjie Chen and Chengyi Wang and Hongli Yu and Yuxuan Song and Xiangpeng Wei and Hao Zhou and Jingjing Liu and Wei-Ying Ma and Ya-Qin Zhang and Lin Yan and Yonghui Wu and Mingxuan Wang},
booktitle={The Thirty-ninth Annual Conference on Neural Information Processing Systems},
year={2026},
url={https://openreview.net/forum?id=2a36EMSSTp}
}

@misc{swift,
      title={SWIFT:A Scalable lightWeight Infrastructure for Fine-Tuning},
      author={Yuze Zhao and Jintao Huang and Jinghan Hu and Xingjun Wang and Yunlin Mao and Daoze Zhang and Zeyinzi Jiang and Zhikai Wu and Baole Ai and Ang Wang and Wenmeng Zhou and Yingda Chen},
      year={2024},
      eprint={2408.05517},
      archivePrefix={arXiv},
      primaryClass={cs.CL},
      url={https://arxiv.org/abs/2408.05517},
}

@inproceedings{
adamw,
title={Decoupled Weight Decay Regularization},
author={Ilya Loshchilov and Frank Hutter},
booktitle={International Conference on Learning Representations},
year={2019},
url={https://openreview.net/forum?id=Bkg6RiCqY7},
}

@inproceedings{8bit,
  title={8-bit Optimizers via Block-wise Quantization},
  author={Dettmers, Tim and Lewis, Mike and Shleifer, Sam and Zettlemoyer, Luke},
  booktitle={International Conference on Learning Representations}
}

@inproceedings{vllm,
  title={Efficient Memory Management for Large Language Model Serving with PagedAttention},
  author={Woosuk Kwon and Zhuohan Li and Siyuan Zhuang and Ying Sheng and Lianmin Zheng and Cody Hao Yu and Joseph E. Gonzalez and Hao Zhang and Ion Stoica},
  booktitle={Proceedings of the ACM SIGOPS 29th Symposium on Operating Systems Principles},
  year={2023}
}

@inproceedings{fa2,
  title={Flash{A}ttention-2: Faster Attention with Better Parallelism and Work Partitioning},
  author={Dao, Tri},
  booktitle={International Conference on Learning Representations (ICLR)},
  year={2024}
}

@misc{padding,
      title={Enhancing Training Efficiency Using Packing with Flash Attention}, 
      author={Achintya Kundu and Rhui Dih Lee and Laura Wynter and Raghu Kiran Ganti and Mayank Mishra},
      year={2024},
      eprint={2407.09105},
      archivePrefix={arXiv},
      primaryClass={cs.LG},
      url={https://arxiv.org/abs/2407.09105}, 
}

@inproceedings{gemba-mqm,
    title = {GEMBA-MQM: Detecting Translation Quality Error Spans with GPT-4},
    author = {Kocmi, Tom  and Federmann, Christian},
    booktitle = "Proceedings of the Eighth Conference on Machine Translation",
    month = dec,
    year = "2023",
    address = "Singapore",
    publisher = "Association for Computational Linguistics",
}

@article{gemini,
  title={Gemini: a family of highly capable multimodal models},
  author={Team, Gemini and Anil, Rohan and Borgeaud, Sebastian and Alayrac, Jean-Baptiste and Yu, Jiahui and Soricut, Radu and Schalkwyk, Johan and Dai, Andrew M and Hauth, Anja and Millican, Katie and others},
  journal={arXiv preprint arXiv:2312.11805},
  year={2023}
}

@inproceedings{metric_bias,
  title={Mitigating Metric Bias in Minimum Bayes Risk Decoding},
  author={Kovacs, Geza and Deutsch, Daniel and Freitag, Markus},
  booktitle={Proceedings of the Ninth Conference on Machine Translation},
  pages={1063--1094},
  year={2024}
}

@inproceedings{wmtm23,
    title = "Results of {WMT}23 Metrics Shared Task: Metrics Might Be Guilty but References Are Not Innocent",
    author = "Freitag, Markus  and
      Mathur, Nitika  and
      Lo, Chi-kiu  and
      Avramidis, Eleftherios  and
      Rei, Ricardo  and
      Thompson, Brian  and
      Kocmi, Tom  and
      Blain, Frederic  and
      Deutsch, Daniel  and
      Stewart, Craig  and
      Zerva, Chrysoula  and
      Castilho, Sheila  and
      Lavie, Alon  and
      Foster, George",
    editor = "Koehn, Philipp  and
      Haddow, Barry  and
      Kocmi, Tom  and
      Monz, Christof",
    booktitle = "Proceedings of the Eighth Conference on Machine Translation",
    month = dec,
    year = "2023",
    address = "Singapore",
    publisher = "Association for Computational Linguistics",
    url = "https://aclanthology.org/2023.wmt-1.51/",
    doi = "10.18653/v1/2023.wmt-1.51",
    pages = "578--628"
}

@inproceedings{wmtm24,
    title = "Are {LLM}s Breaking {MT} Metrics? Results of the {WMT}24 Metrics Shared Task",
    author = "Freitag, Markus  and
      Mathur, Nitika  and
      Deutsch, Daniel  and
      Lo, Chi-Kiu  and
      Avramidis, Eleftherios  and
      Rei, Ricardo  and
      Thompson, Brian  and
      Blain, Frederic  and
      Kocmi, Tom  and
      Wang, Jiayi  and
      Adelani, David Ifeoluwa  and
      Buchicchio, Marianna  and
      Zerva, Chrysoula  and
      Lavie, Alon",
    editor = "Haddow, Barry  and
      Kocmi, Tom  and
      Koehn, Philipp  and
      Monz, Christof",
    booktitle = "Proceedings of the Ninth Conference on Machine Translation",
    month = nov,
    year = "2024",
    address = "Miami, Florida, USA",
    publisher = "Association for Computational Linguistics",
    url = "https://aclanthology.org/2024.wmt-1.2/",
    doi = "10.18653/v1/2024.wmt-1.2",
    pages = "47--81"
}

@inproceedings{wmt19,
    title = "Findings of the 2019 Conference on Machine Translation ({WMT}19)",
    author = {Barrault, Lo{\"i}c  and
      Bojar, Ond{\v{r}}ej  and
      Costa-juss{\`a}, Marta R.  and
      Federmann, Christian  and
      Fishel, Mark  and
      Graham, Yvette  and
      Haddow, Barry  and
      Huck, Matthias  and
      Koehn, Philipp  and
      Malmasi, Shervin  and
      Monz, Christof  and
      M{\"u}ller, Mathias  and
      Pal, Santanu  and
      Post, Matt  and
      Zampieri, Marcos},
    editor = "Bojar, Ond{\v{r}}ej  and
      Chatterjee, Rajen  and
      Federmann, Christian  and
      Fishel, Mark  and
      Graham, Yvette  and
      Haddow, Barry  and
      Huck, Matthias  and
      Yepes, Antonio Jimeno  and
      Koehn, Philipp  and
      Martins, Andr{\'e}  and
      Monz, Christof  and
      Negri, Matteo  and
      N{\'e}v{\'e}ol, Aur{\'e}lie  and
      Neves, Mariana  and
      Post, Matt  and
      Turchi, Marco  and
      Verspoor, Karin",
    booktitle = "Proceedings of the Fourth Conference on Machine Translation (Volume 2: Shared Task Papers, Day 1)",
    month = aug,
    year = "2019",
    address = "Florence, Italy",
    publisher = "Association for Computational Linguistics",
    url = "https://aclanthology.org/W19-5301/",
    doi = "10.18653/v1/W19-5301",
    pages = "1--61"
}

@inproceedings{wmt20,
    title = "Findings of the 2020 Conference on Machine Translation ({WMT}20)",
    author = {Barrault, Lo{\"i}c  and
      Biesialska, Magdalena  and
      Bojar, Ond{\v{r}}ej  and
      Costa-juss{\`a}, Marta R.  and
      Federmann, Christian  and
      Graham, Yvette  and
      Grundkiewicz, Roman  and
      Haddow, Barry  and
      Huck, Matthias  and
      Joanis, Eric  and
      Kocmi, Tom  and
      Koehn, Philipp  and
      Lo, Chi-kiu  and
      Ljube{\v{s}}i{\'c}, Nikola  and
      Monz, Christof  and
      Morishita, Makoto  and
      Nagata, Masaaki  and
      Nakazawa, Toshiaki  and
      Pal, Santanu  and
      Post, Matt  and
      Zampieri, Marcos},
    editor = {Barrault, Lo{\"i}c  and
      Bojar, Ond{\v{r}}ej  and
      Bougares, Fethi  and
      Chatterjee, Rajen  and
      Costa-juss{\`a}, Marta R.  and
      Federmann, Christian  and
      Fishel, Mark  and
      Fraser, Alexander  and
      Graham, Yvette  and
      Guzman, Paco  and
      Haddow, Barry  and
      Huck, Matthias  and
      Yepes, Antonio Jimeno  and
      Koehn, Philipp  and
      Martins, Andr{\'e}  and
      Morishita, Makoto  and
      Monz, Christof  and
      Nagata, Masaaki  and
      Nakazawa, Toshiaki  and
      Negri, Matteo},
    booktitle = "Proceedings of the Fifth Conference on Machine Translation",
    month = nov,
    year = "2020",
    address = "Online",
    publisher = "Association for Computational Linguistics",
    url = "https://aclanthology.org/2020.wmt-1.1/",
    doi = "10.18653/v1/2020.wmt-1.1",
    pages = "1--55"
}

@inproceedings{wmt21,
    title = "Findings of the 2021 Conference on Machine Translation ({WMT}21)",
    author = "Akhbardeh, Farhad  and
      Arkhangorodsky, Arkady  and
      Biesialska, Magdalena  and
      Bojar, Ond{\v{r}}ej  and
      Chatterjee, Rajen  and
      Chaudhary, Vishrav  and
      Costa-jussa, Marta R.  and
      Espa{\~n}a-Bonet, Cristina  and
      Fan, Angela  and
      Federmann, Christian  and
      Freitag, Markus  and
      Graham, Yvette  and
      Grundkiewicz, Roman  and
      Haddow, Barry  and
      Harter, Leonie  and
      Heafield, Kenneth  and
      Homan, Christopher  and
      Huck, Matthias  and
      Amponsah-Kaakyire, Kwabena  and
      Kasai, Jungo  and
      Khashabi, Daniel  and
      Knight, Kevin  and
      Kocmi, Tom  and
      Koehn, Philipp  and
      Lourie, Nicholas  and
      Monz, Christof  and
      Morishita, Makoto  and
      Nagata, Masaaki  and
      Nagesh, Ajay  and
      Nakazawa, Toshiaki  and
      Negri, Matteo  and
      Pal, Santanu  and
      Tapo, Allahsera Auguste  and
      Turchi, Marco  and
      Vydrin, Valentin  and
      Zampieri, Marcos",
    editor = "Barrault, Loic  and
      Bojar, Ondrej  and
      Bougares, Fethi  and
      Chatterjee, Rajen  and
      Costa-jussa, Marta R.  and
      Federmann, Christian  and
      Fishel, Mark  and
      Fraser, Alexander  and
      Freitag, Markus  and
      Graham, Yvette  and
      Grundkiewicz, Roman  and
      Guzman, Paco  and
      Haddow, Barry  and
      Huck, Matthias  and
      Yepes, Antonio Jimeno  and
      Koehn, Philipp  and
      Kocmi, Tom  and
      Martins, Andre  and
      Morishita, Makoto  and
      Monz, Christof",
    booktitle = "Proceedings of the Sixth Conference on Machine Translation",
    month = nov,
    year = "2021",
    address = "Online",
    publisher = "Association for Computational Linguistics",
    url = "https://aclanthology.org/2021.wmt-1.1/",
    pages = "1--88"
}

@inproceedings{wmt22,
    title = "Findings of the 2022 Conference on Machine Translation ({WMT}22)",
    author = "Kocmi, Tom  and
      Bawden, Rachel  and
      Bojar, Ond{\v{r}}ej  and
      Dvorkovich, Anton  and
      Federmann, Christian  and
      Fishel, Mark  and
      Gowda, Thamme  and
      Graham, Yvette  and
      Grundkiewicz, Roman  and
      Haddow, Barry  and
      Knowles, Rebecca  and
      Koehn, Philipp  and
      Monz, Christof  and
      Morishita, Makoto  and
      Nagata, Masaaki  and
      Nakazawa, Toshiaki  and
      Nov{\'a}k, Michal  and
      Popel, Martin  and
      Popovi{\'c}, Maja",
    editor = {Koehn, Philipp  and
      Barrault, Lo{\"i}c  and
      Bojar, Ond{\v{r}}ej  and
      Bougares, Fethi  and
      Chatterjee, Rajen  and
      Costa-juss{\`a}, Marta R.  and
      Federmann, Christian  and
      Fishel, Mark  and
      Fraser, Alexander  and
      Freitag, Markus  and
      Graham, Yvette  and
      Grundkiewicz, Roman  and
      Guzman, Paco  and
      Haddow, Barry  and
      Huck, Matthias  and
      Jimeno Yepes, Antonio  and
      Kocmi, Tom  and
      Martins, Andr{\'e}  and
      Morishita, Makoto  and
      Monz, Christof  and
      Nagata, Masaaki  and
      Nakazawa, Toshiaki  and
      Negri, Matteo  and
      N{\'e}v{\'e}ol, Aur{\'e}lie  and
      Neves, Mariana  and
      Popel, Martin  and
      Turchi, Marco  and
      Zampieri, Marcos},
    booktitle = "Proceedings of the Seventh Conference on Machine Translation (WMT)",
    month = dec,
    year = "2022",
    address = "Abu Dhabi, United Arab Emirates (Hybrid)",
    publisher = "Association for Computational Linguistics",
    url = "https://aclanthology.org/2022.wmt-1.1/",
    doi = "10.18653/v1/2022.wmt-1.1",
    pages = "1--45"
}

@inproceedings{wmt23,
    title = "Findings of the 2023 Conference on Machine Translation ({WMT}23): {LLM}s Are Here but Not Quite There Yet",
    author = "Kocmi, Tom  and
      Avramidis, Eleftherios  and
      Bawden, Rachel  and
      Bojar, Ond{\v{r}}ej  and
      Dvorkovich, Anton  and
      Federmann, Christian  and
      Fishel, Mark  and
      Freitag, Markus  and
      Gowda, Thamme  and
      Grundkiewicz, Roman  and
      Haddow, Barry  and
      Koehn, Philipp  and
      Marie, Benjamin  and
      Monz, Christof  and
      Morishita, Makoto  and
      Murray, Kenton  and
      Nagata, Masaaki  and
      Nakazawa, Toshiaki  and
      Popel, Martin  and
      Popovi{\'c}, Maja  and
      Shmatova, Mariya  and
      Suzuki, Jun",
    editor = "Koehn, Philipp  and
      Haddow, Barry  and
      Kocmi, Tom  and
      Monz, Christof",
    booktitle = "Proceedings of the Eighth Conference on Machine Translation",
    month = dec,
    year = "2023",
    address = "Singapore",
    publisher = "Association for Computational Linguistics",
    url = "https://aclanthology.org/2023.wmt-1.1/",
    doi = "10.18653/v1/2023.wmt-1.1",
    pages = "1--42"
}

@inproceedings{wmt24,
    title = "Findings of the {WMT}24 General Machine Translation Shared Task: The {LLM} Era Is Here but {MT} Is Not Solved Yet",
    author = "Kocmi, Tom  and
      Avramidis, Eleftherios  and
      Bawden, Rachel  and
      Bojar, Ond{\v{r}}ej  and
      Dvorkovich, Anton  and
      Federmann, Christian  and
      Fishel, Mark  and
      Freitag, Markus  and
      Gowda, Thamme  and
      Grundkiewicz, Roman  and
      Haddow, Barry  and
      Karpinska, Marzena  and
      Koehn, Philipp  and
      Marie, Benjamin  and
      Monz, Christof  and
      Murray, Kenton  and
      Nagata, Masaaki  and
      Popel, Martin  and
      Popovi{\'c}, Maja  and
      Shmatova, Mariya  and
      Steingr{\'i}msson, Steinth{\'o}r  and
      Zouhar, Vil{\'e}m",
    editor = "Haddow, Barry  and
      Kocmi, Tom  and
      Koehn, Philipp  and
      Monz, Christof",
    booktitle = "Proceedings of the Ninth Conference on Machine Translation",
    month = nov,
    year = "2024",
    address = "Miami, Florida, USA",
    publisher = "Association for Computational Linguistics",
    url = "https://aclanthology.org/2024.wmt-1.1/",
    doi = "10.18653/v1/2024.wmt-1.1",
    pages = "1--46"
}

@inproceedings{pbs,
    title = "Statistical Significance Tests for Machine Translation Evaluation",
    author = "Koehn, Philipp",
    editor = "Lin, Dekang  and
      Wu, Dekai",
    booktitle = "Proceedings of the 2004 Conference on Empirical Methods in Natural Language Processing",
    month = jul,
    year = "2004",
    address = "Barcelona, Spain",
    publisher = "Association for Computational Linguistics",
    url = "https://aclanthology.org/W04-3250/",
    pages = "388--395"
}

@inproceedings{ape1,
    title = "Effects of Automatic Rewriting of Source Language within a {J}apanese to {E}nglish {MT} System",
    author = "Shirai, Satoshi  and
      Ikehara, Satoru  and
      Kawaoka, Tsukasa",
    booktitle = "Proceedings of the Fifth Conference on Theoretical and Methodological Issues in Machine Translation of Natural Languages",
    month = jul # " 14-16",
    year = "1993",
    address = "Kyoto, Japan",
    url = "https://aclanthology.org/1993.tmi-1.20/"
}

@inproceedings{ape2,
  title={Automatic rewriting for controlled language translation},
  author={Mitamura, Teruko and Nyberg, Eric}
}

@inproceedings{ape3,
    title = "Improvement of translation quality of {E}nglish newspaper headlines by automatic preediting",
    author = "Yoshimi, Takehiko  and
      Sata, Ichiko",
    booktitle = "Proceedings of Machine Translation Summit VII",
    month = sep # " 13-17",
    year = "1999",
    address = "Singapore, Singapore",
    url = "https://aclanthology.org/1999.mtsummit-1.73/",
    pages = "496--500"
}

@inproceedings{reorder1,
    title = "Improving a Statistical {MT} System with Automatically Learned Rewrite Patterns",
    author = "Xia, Fei  and
      McCord, Michael",
    booktitle = "{COLING} 2004: Proceedings of the 20th International Conference on Computational Linguistics",
    month = "aug 23–aug 27",
    year = "2004",
    address = "Geneva, Switzerland",
    publisher = "COLING",
    url = "https://aclanthology.org/C04-1073/",
    pages = "508--514"
}

@inproceedings{reorder2,
    title = "A Probabilistic Approach to Syntax-based Reordering for Statistical Machine Translation",
    author = "Li, Chi-Ho  and
      Li, Minghui  and
      Zhang, Dongdong  and
      Li, Mu  and
      Zhou, Ming  and
      Guan, Yi",
    editor = "Zaenen, Annie  and
      van den Bosch, Antal",
    booktitle = "Proceedings of the 45th Annual Meeting of the Association of Computational Linguistics",
    month = jun,
    year = "2007",
    address = "Prague, Czech Republic",
    publisher = "Association for Computational Linguistics",
    url = "https://aclanthology.org/P07-1091/",
    pages = "720--727"
}

@inproceedings{reorder3,
    title = "Discriminative Preordering Meets Kendall{'}s $\tau$ Maximization",
    author = "Hoshino, Sho  and
      Miyao, Yusuke  and
      Sudoh, Katsuhito  and
      Hayashi, Katsuhiko  and
      Nagata, Masaaki",
    editor = "Zong, Chengqing  and
      Strube, Michael",
    booktitle = "Proceedings of the 53rd Annual Meeting of the Association for Computational Linguistics and the 7th International Joint Conference on Natural Language Processing (Volume 2: Short Papers)",
    month = jul,
    year = "2015",
    address = "Beijing, China",
    publisher = "Association for Computational Linguistics",
    url = "https://aclanthology.org/P15-2023/",
    doi = "10.3115/v1/P15-2023",
    pages = "139--144"
}

@inproceedings{ape_eval1,
    title = "Evaluating Neural Machine Translation in {E}nglish-{J}apanese Task",
    author = "Zhu, Zhongyuan",
    editor = "Nakazawa, Toshiaki  and
      Mino, Hideya  and
      Goto, Isao  and
      Neubig, Graham  and
      Kurohashi, Sadao  and
      Sumita, Eiichiro",
    booktitle = "Proceedings of the 2nd Workshop on {A}sian Translation ({WAT}2015)",
    month = oct,
    year = "2015",
    address = "Kyoto, Japan",
    publisher = "Workshop on Asian Translation",
    url = "https://aclanthology.org/W15-5007/",
    pages = "61--68"
}

@article{ape_eval2,
author = {Du, Jinhua and Way, Andy},
year = {2017},
month = {06},
pages = {},
title = {Pre-Reordering for Neural Machine Translation: Helpful or Harmful?},
volume = {108},
journal = {The Prague Bulletin of Mathematical Linguistics},
doi = {10.1515/pralin-2017-0018}
}

@inproceedings{fujita17,
author = {Miyata, Rei and Fujita, Atsushi},
year = {2017},
month = {05},
pages = {},
title = {Dissecting Human Pre-Editing toward Better Use of Off-the-Shelf Machine Translation Systems}
}

@inproceedings{gpt4,
  title={GPT-4 Technical Report},
  author={OpenAI Josh Achiam and Steven Adler and Sandhini Agarwal and Lama Ahmad and Ilge Akkaya and Florencia Leoni Aleman and Diogo Almeida and Janko Altenschmidt and Sam Altman and Shyamal Anadkat and Red Avila and Igor Babuschkin and Suchir Balaji and Valerie Balcom and Paul Baltescu and Haiming Bao and Mo Bavarian and Jeff Belgum and Irwan Bello and others},
  year={2023},
  url={https://api.semanticscholar.org/CorpusID:257532815}
}

@misc{gpt3,
      title={Language Models are Few-Shot Learners}, 
      author={Tom B. Brown and Benjamin Mann and Nick Ryder and Melanie Subbiah and Jared Kaplan and Prafulla Dhariwal and Arvind Neelakantan and Pranav Shyam and Girish Sastry and Amanda Askell and Sandhini Agarwal and Ariel Herbert-Voss and Gretchen Krueger and Tom Henighan and Rewon Child and Aditya Ramesh and Daniel M. Ziegler and Jeffrey Wu and Clemens Winter and Christopher Hesse and Mark Chen and Eric Sigler and Mateusz Litwin and Scott Gray and Benjamin Chess and Jack Clark and Christopher Berner and Sam McCandlish and Alec Radford and Ilya Sutskever and Dario Amodei},
      year={2020},
      eprint={2005.14165},
      archivePrefix={arXiv},
      primaryClass={cs.CL},
      url={https://arxiv.org/abs/2005.14165}, 
}

@misc{
token_sft,
title={Learning from others' mistakes: Finetuning machine translation models with span-level error annotations},
author={Lily H Zhang and Hamid Dadkhahi and Mara Finkelstein and Firas Trabelsi and Jiaming Luo and Markus Freitag},
year={2025},
url={https://openreview.net/forum?id=204sPiwBbB}
}

@misc{gemma3,
      title={Gemma 3 Technical Report}, 
      author={Gemma Team and Aishwarya Kamath and Johan Ferret and Shreya Pathak and Nino Vieillard and Ramona Merhej and Sarah Perrin and Tatiana Matejovicova and Alexandre Ramé and Morgane Rivière and Louis Rouillard and Thomas Mesnard and Geoffrey Cideron and Jean-bastien Grill and Sabela Ramos and Edouard Yvinec and Michelle Casbon and Etienne Pot and Ivo Penchev and Gaël Liu and Francesco Visin and Kathleen Kenealy and Lucas Beyer and Xiaohai Zhai and Anton Tsitsulin and Robert Busa-Fekete and Alex Feng and Noveen Sachdeva and Benjamin Coleman and Yi Gao and Basil Mustafa and Iain Barr and Emilio Parisotto and David Tian and Matan Eyal and Colin Cherry and Jan-Thorsten Peter and Danila Sinopalnikov and Surya Bhupatiraju and Rishabh Agarwal and Mehran Kazemi and Dan Malkin and Ravin Kumar and David Vilar and Idan Brusilovsky and Jiaming Luo and Andreas Steiner and Abe Friesen and Abhanshu Sharma and Abheesht Sharma and Adi Mayrav Gilady and Adrian Goedeckemeyer and Alaa Saade and Alex Feng and Alexander Kolesnikov and Alexei Bendebury and Alvin Abdagic and Amit Vadi and András György and André Susano Pinto and Anil Das and Ankur Bapna and Antoine Miech and Antoine Yang and Antonia Paterson and Ashish Shenoy and Ayan Chakrabarti and Bilal Piot and Bo Wu and Bobak Shahriari and Bryce Petrini and Charlie Chen and Charline Le Lan and Christopher A. Choquette-Choo and CJ Carey and Cormac Brick and Daniel Deutsch and Danielle Eisenbud and Dee Cattle and Derek Cheng and Dimitris Paparas and Divyashree Shivakumar Sreepathihalli and Doug Reid and Dustin Tran and Dustin Zelle and Eric Noland and Erwin Huizenga and Eugene Kharitonov and Frederick Liu and Gagik Amirkhanyan and Glenn Cameron and Hadi Hashemi and Hanna Klimczak-Plucińska and Harman Singh and Harsh Mehta and Harshal Tushar Lehri and Hussein Hazimeh and Ian Ballantyne and Idan Szpektor and Ivan Nardini and Jean Pouget-Abadie and Jetha Chan and Joe Stanton and John Wieting and Jonathan Lai and Jordi Orbay and Joseph Fernandez and Josh Newlan and Ju-yeong Ji and Jyotinder Singh and Kat Black and Kathy Yu and Kevin Hui and Kiran Vodrahalli and Klaus Greff and Linhai Qiu and Marcella Valentine and Marina Coelho and Marvin Ritter and Matt Hoffman and Matthew Watson and Mayank Chaturvedi and Michael Moynihan and Min Ma and Nabila Babar and Natasha Noy and Nathan Byrd and Nick Roy and Nikola Momchev and Nilay Chauhan and Noveen Sachdeva and Oskar Bunyan and Pankil Botarda and Paul Caron and Paul Kishan Rubenstein and Phil Culliton and Philipp Schmid and Pier Giuseppe Sessa and Pingmei Xu and Piotr Stanczyk and Pouya Tafti and Rakesh Shivanna and Renjie Wu and Renke Pan and Reza Rokni and Rob Willoughby and Rohith Vallu and Ryan Mullins and Sammy Jerome and Sara Smoot and Sertan Girgin and Shariq Iqbal and Shashir Reddy and Shruti Sheth and Siim Põder and Sijal Bhatnagar and Sindhu Raghuram Panyam and Sivan Eiger and Susan Zhang and Tianqi Liu and Trevor Yacovone and Tyler Liechty and Uday Kalra and Utku Evci and Vedant Misra and Vincent Roseberry and Vlad Feinberg and Vlad Kolesnikov and Woohyun Han and Woosuk Kwon and Xi Chen and Yinlam Chow and Yuvein Zhu and Zichuan Wei and Zoltan Egyed and Victor Cotruta and Minh Giang and Phoebe Kirk and Anand Rao and Kat Black and Nabila Babar and Jessica Lo and Erica Moreira and Luiz Gustavo Martins and Omar Sanseviero and Lucas Gonzalez and Zach Gleicher and Tris Warkentin and Vahab Mirrokni and Evan Senter and Eli Collins and Joelle Barral and Zoubin Ghahramani and Raia Hadsell and Yossi Matias and D. Sculley and Slav Petrov and Noah Fiedel and Noam Shazeer and Oriol Vinyals and Jeff Dean and Demis Hassabis and Koray Kavukcuoglu and Clement Farabet and Elena Buchatskaya and Jean-Baptiste Alayrac and Rohan Anil and Dmitry and Lepikhin and Sebastian Borgeaud and Olivier Bachem and Armand Joulin and Alek Andreev and Cassidy Hardin and Robert Dadashi and Léonard Hussenot},
      year={2025},
      eprint={2503.19786},
      archivePrefix={arXiv},
      primaryClass={cs.CL},
      url={https://arxiv.org/abs/2503.19786}, 
}

@misc{qwen3, 
      title={Qwen3 Technical Report}, 
      author={An Yang and Anfeng Li and Baosong Yang and Beichen Zhang and Binyuan Hui and Bo Zheng and Bowen Yu and Chang Gao and Chengen Huang and Chenxu Lv and Chujie Zheng and Dayiheng Liu and Fan Zhou and Fei Huang and Feng Hu and Hao Ge and Haoran Wei and Huan Lin and Jialong Tang and Jian Yang and Jianhong Tu and Jianwei Zhang and Jianxin Yang and Jiaxi Yang and Jing Zhou and Jingren Zhou and Junyang Lin and Kai Dang and Keqin Bao and Kexin Yang and Le Yu and Lianghao Deng and Mei Li and Mingfeng Xue and Mingze Li and Pei Zhang and Peng Wang and Qin Zhu and Rui Men and Ruize Gao and Shixuan Liu and Shuang Luo and Tianhao Li and Tianyi Tang and Wenbiao Yin and Xingzhang Ren and Xinyu Wang and Xinyu Zhang and Xuancheng Ren and Yang Fan and Yang Su and Yichang Zhang and Yinger Zhang and Yu Wan and Yuqiong Liu and Zekun Wang and Zeyu Cui and Zhenru Zhang and Zhipeng Zhou and Zihan Qiu},
      year={2025},
      eprint={2505.09388},
      archivePrefix={arXiv},
      primaryClass={cs.CL},
      url={https://arxiv.org/abs/2505.09388}, 
}

@misc{translategemma,
      title={TranslateGemma Technical Report}, 
      author={Mara Finkelstein and Isaac Caswell and Tobias Domhan and Jan-Thorsten Peter and Juraj Juraska and Parker Riley and Daniel Deutsch and Geza Kovacs and Cole Dilanni and Colin Cherry and Eleftheria Briakou and Elizabeth Nielsen and Jiaming Luo and Kat Black and Ryan Mullins and Sweta Agrawal and Wenda Xu and Erin Kats and Stephane Jaskiewicz and Markus Freitag and David Vilar},
      year={2026},
      eprint={2601.09012},
      archivePrefix={arXiv},
      primaryClass={cs.CL},
      url={https://arxiv.org/abs/2601.09012}, 
}

@article{xcomet,
    title = "x{COMET}: Transparent Machine Translation Evaluation through Fine-grained Error Detection",
    author = "Guerreiro, Nuno M.  and
      Rei, Ricardo  and
      Stigt, Daan van  and
      Coheur, Luisa  and
      Colombo, Pierre  and
      Martins, Andr{\'e} F. T.",
    journal = "Transactions of the Association for Computational Linguistics",
    volume = "12",
    year = "2024",
    address = "Cambridge, MA",
    publisher = "MIT Press",
    url = "https://aclanthology.org/2024.tacl-1.54/",
    doi = "10.1162/tacl_a_00683",
    pages = "979--995"
}

@inproceedings{metricx24,
    title = "{M}etric{X}-24: The {G}oogle Submission to the {WMT} 2024 Metrics Shared Task",
    author = "Juraska, Juraj  and
      Deutsch, Daniel  and
      Finkelstein, Mara  and
      Freitag, Markus",
    editor = "Haddow, Barry  and
      Kocmi, Tom  and
      Koehn, Philipp  and
      Monz, Christof",
    booktitle = "Proceedings of the Ninth Conference on Machine Translation",
    month = nov,
    year = "2024",
    address = "Miami, Florida, USA",
    publisher = "Association for Computational Linguistics",
    url = "https://aclanthology.org/2024.wmt-1.35/",
    doi = "10.18653/v1/2024.wmt-1.35",
    pages = "492--504"
}

@inproceedings{
dpo,
title={Direct Preference Optimization: Your Language Model is Secretly a Reward Model},
author={Rafael Rafailov and Archit Sharma and Eric Mitchell and Christopher D Manning and Stefano Ermon and Chelsea Finn},
booktitle={Thirty-seventh Conference on Neural Information Processing Systems},
year={2023},
url={https://openreview.net/forum?id=HPuSIXJaa9}
}

@inproceedings{dare,
author = {Yu, Le and Yu, Bowen and Yu, Haiyang and Huang, Fei and Li, Yongbin},
title = {Language models are super mario: absorbing abilities from homologous models as a free lunch},
year = {2024},
publisher = {JMLR.org},
booktitle = {Proceedings of the 41st International Conference on Machine Learning},
articleno = {2382},
numpages = {21},
location = {Vienna, Austria},
series = {ICML'24}
}

\clearpage
\appendix

\section{Implementation Details of RLSR}
\label{sec:rlsr_details}
Our training codebase was built on \texttt{ms-swift} \cite{swift}. 
RLSR training was conducted on several servers, each equipped with 4$\times$ NVIDIA H100 GPUs. 
We trained the model for at most one epoch with an effective batch size of 512 (accounting for gradient accumulation and multi-GPU parallelism). 
We used the 8-bit AdamW optimizer \cite{adamw,8bit} with a maximum learning rate of $1 \times 10^{-6}$ and a cosine decay learning rate scheduler. 
For the DAPO hyperparameters, we set the number of sampled responses per prompt to 64, the sampling temperature to 1.5, and the coefficient $\beta$ of the KL penalty in Eq.~\ref{eq:total_reward} to 0.04.
During training, response sampling was performed with \texttt{vLLM} \cite{vllm}; padding-free computations \cite{padding} and Flash Attention \cite{fa2} were used to accelerate training.

\rev{\section{Model Merging with DARE}
\label{sec:dare_details}

This appendix gives the detailed mathematical definition of the merging procedure used to construct \textbf{RLSR-Ens} (Section~\ref{sec:ensemble}).
We merge $K$ specialist rewriting models, each trained with RLSR for one MT model $M_k$ and all fine-tuned from the same base model $R_{\text{ref}}$, using DARE~\cite{dare}.

\paragraph{Task vectors}
Let $\theta_{\text{ref}} \in \mathbb{R}^d$ denote the parameters of the base model $R_{\text{ref}}$, and let $\theta_k \in \mathbb{R}^d$ denote the parameters of the $k$-th specialist $R_{\theta_k}$ obtained by RLSR fine-tuning $R_{\text{ref}}$ for $M_k$.
Following task arithmetic, we work with the \emph{task vector} of each specialist, defined as the element-wise difference between the fine-tuned and base parameters,
\begin{equation}
    \delta_k = \theta_k - \theta_{\text{ref}}, \qquad k = 1, \dots, K.
    \label{eq:task_vector}
\end{equation}
DARE operates on the task vectors $\delta_k$ rather than on the raw weights $\theta_k$, and adds the processed vectors back to the shared base $\theta_{\text{ref}}$.

\paragraph{Drop and rescale}
DARE sparsifies each task vector by randomly dropping a fraction of its entries and rescaling the survivors.
For every specialist $k$, we sample an element-wise Bernoulli mask $m_k \in \{0,1\}^d$ with
\begin{equation}
    m_{k,i} \sim \text{Bernoulli}(p), \qquad i = 1, \dots, d,
    \label{eq:dare_mask}
\end{equation}
where $p$ is the \emph{drop rate}: $m_{k,i} = 1$ indicates that the $i$-th coordinate of $\delta_k$ is dropped, so each entry survives with probability $1 - p$.
The drop-and-rescale operation then yields the DARE-processed task vector:
\begin{equation}
    \hat{\delta}_k = \frac{1}{1 - p}\,\big(\mathbf{1} - m_k\big) \odot \delta_k,
    \label{eq:dare_drop}
\end{equation}
where $\odot$ is the element-wise (Hadamard) product and $\mathbf{1}$ is the all-ones vector.
The factor $1/(1 - p)$ is the inverted-dropout unbiasing term: it keeps each processed entry unbiased in expectation, since:
\begin{align}
    \mathbb{E}\big[\hat{\delta}_{k,i}\big]
    &= \frac{1}{1 - p}\,\mathbb{E}\big[1 - m_{k,i}\big]\,\delta_{k,i} \nonumber\\
    &= \frac{1 - p}{1 - p}\,\delta_{k,i}
    = \delta_{k,i},
    \label{eq:dare_expectation}
\end{align}
so that $\mathbb{E}[\hat{\delta}_k] = \delta_k$.
Rescaling thus compensates for the $p$ fraction of coordinates zeroed by the mask, preserving the expected magnitude of the task vector after sparsification.

\paragraph{Equal-weight merge}
We combine the DARE-processed task vectors by weighting all $K$ specialists equally, i.e., by averaging them and adding the result back to the base model.
The parameters of RLSR-Ens are:
\begin{equation}
    \theta_{\text{Ens}}
    = \theta_{\text{ref}} + \frac{1}{K} \sum_{k=1}^{K} \hat{\delta}_k .
    \label{eq:dare_merge}
\end{equation}
The merged model $R_{\theta_{\text{Ens}}}$ requires no further training and no access to the MT models.

\paragraph{Hyperparameters}
We set the task-vector density to $0.5$, i.e., the drop rate $p = 0.5$ in Eq.~\ref{eq:dare_drop}, so that half of each task vector's entries are dropped in expectation and the survivors are rescaled by a factor $1/(1 - p) = 2$, as recommended by \citet{dare}.
Here, \emph{density} denotes the fraction of task-vector entries kept ($\text{density} = 1 - p$), following common merging tooling.}

\section{Implementation Details of Qwen3 4B SFT-RL}
\label{sec:qwen_sft_rl}

For both the SFT and RL stages, we used the same source--reference translation pairs used to train the rewriting models. 
We randomly selected 5\% of the data as a validation set and used the remainder for training. The training codebase was also built on \texttt{ms-swift} \cite{swift}.

\paragraph{SFT Stage}
We set the effective batch size to 256 and trained for at most three epochs. We used a learning rate of $2 \times 10^{-5}$ with a cosine decay scheduler and a warmup ratio of 0.05, and adopted the 8-bit AdamW optimizer.

\paragraph{RL Stage}
We employed the DAPO algorithm \cite{dapo}, initializing from either the original Qwen3 4B or the SFT checkpoint described above. We set the effective batch size to 256, trained for at most one epoch, and used a learning rate of $1 \times 10^{-6}$ with cosine decay and a warmup ratio of 0.05. We again used the 8-bit AdamW optimizer. We set $\epsilon_{\text{high}}$ to 0.28 and the KL divergence coefficient to 0.04. For each sample, we generated 64 outputs and used \texttt{xCOMET-XL} as the reward model.

\paragraph{Effect of Training Strategies}
We evaluated how different training strategies affected the resulting MT model: no training (Qwen3 4B), SFT only (Qwen3 4B SFT), RL only (Qwen3 4B RL), and SFT followed by RL (Qwen3 4B SFT-RL). As shown in Table~\ref{tab:exp_qwen_mt}, RL-based training yields substantial gains over the untrained base model across all metrics, while SFT alone provides little benefit and even degrades xCOMET. Combining the two stages (SFT-RL) yields the best overall performance, which is why we adopted this checkpoint as one of our downstream MT models.

\begin{table}[h]
\centering
\setlength\tabcolsep{2pt}
\small
\begin{tabular}{lccccc}
\toprule
Training & xCOMET$\uparrow$ & MetricX$\downarrow$ & GEMBA$\downarrow$ \\
\midrule
-  & 32.88 & 12.37 & 22.81  \\
SFT & 31.37 & 12.22 & 21.14\\
RL & 35.33 & 11.39 & 20.01\\
SFT-RL & \textbf{36.12} & \textbf{10.95} & \textbf{19.05}\\
\bottomrule
\end{tabular}
\caption{Performance of the Qwen3 4B MT model under different training strategies, evaluated on the WMT2025 General MT Shared Task.}
\label{tab:exp_qwen_mt}
\end{table}

\section{Exploring the Gemma4 Family as the Rewriting Model}
\label{sec:gemma4}
\begin{table*}[t]
\centering
\small
\begin{tabular}{llcccccc}
\toprule
\multirow{2}{*}{Rewriting Model} & \multirow{2}{*}{Rewriting Prompt} & \multicolumn{6}{c}{Downstream MT Model} \\
\cmidrule(lr){3-8}
 & & xCOMET$\uparrow$ & MetricX$\downarrow$ & GEMBA$\downarrow$ & xCOMET$\uparrow$ & MetricX$\downarrow$ & GEMBA$\downarrow$ \\
\midrule
& & \multicolumn{3}{c}{Gemma3 27B} & \multicolumn{3}{c}{Gemma4 31B} \\
\midrule
- & - & 49.22 & 7.19 & \textbf{8.73} & \textbf{51.73}& 6.83&\textbf{6.38} \\
\hdashline
Gemma4 4B & Simplification & 47.88 & 7.32 & 9.77 & 50.16 & 6.96 & 6.80 \\
Gemma4 4B & Paraphrase & 44.11 & 7.24 & 9.93 & 45.16 & 7.06 & 7.84 \\
Gemma4 4B & Easy Translate & 49.05 & 7.07 & 8.91& 50.79 & 6.76 & 6.56 \\
\hdashline
Gemma4 31B & Simplification & \textbf{49.30} & 7.18 & 9.01& 50.68 & 6.97& 6.94\\
Gemma4 31B & Paraphrase & 42.45 & 7.19 & 12.55 &43.52 & 7.09 & 7.49\\
Gemma4 31B & Easy Translate & 48.87 & \textbf{6.79} & 8.97&49.40 & \textbf{6.71}& 6.88\\
\bottomrule
\end{tabular}
\caption{Prompt-based rewriting with the Gemma4 family on the WMT2025 General MT Shared Task. The rewriting models are Gemma4 4B and Gemma4 31B, and the downstream MT models are Gemma3 27B and Gemma4 31B.}
\label{tab:gemma4_exp}
\end{table*}
As noted in Section~\ref{sec:exp_setup}, the experiments in the main text use the Qwen3 family for the rewriting model when studying the effectiveness of the prompt-based rewriting baselines and RLSR.
We also explore the effectiveness of the Gemma4 family, a widely used family of open-weight LLMs, as rewriting models.
As a first step, we evaluated the prompt-based rewriting baselines.
We used two model sizes as rewriting models: Gemma4 4B (gemma-4-E4B-it,\footnote{\url{https://huggingface.co/google/gemma-4-E4B-it}} where ``4B'' denotes its effective parameter count) and Gemma4 31B (gemma-4-31B-it), the largest model in the Gemma4 family.
To assess the effectiveness of the Gemma4 family as rewriting models at a lower experimental cost, we restricted these experiments to two downstream MT models, Gemma3 27B and Gemma4 31B, rather than all six MT models used in the main experiments.

As Table~\ref{tab:gemma4_exp} shows, prompt-based rewriting with the Gemma4 family does not yield consistent score improvements over the no-rewriting baseline across most metrics and MT models.
Given this lack of consistent gains, we did not further investigate RLSR or prompt-based rewriting with the Gemma4 family; we therefore conducted our main experiments with the Qwen3 family as the rewriting model.

\section{Prompts}
\label{sec:prompt}

This section describes the prompts used for the MT and rewriting models. In the templates below, \texttt{\{src\_lang\}}, \texttt{\{tgt\_lang\}}, and \texttt{\{input\_text\}} are placeholders for the source language, the target language, and the input text, respectively.

\paragraph{MT Prompt}
For all MT models except Translategemma 27B (which uses its own model-specific prompt format), we used the following prompt:
\begin{quote}\ttfamily
Please translate the following \{src\_lang\} document into \{tgt\_lang\}.\textbackslash n
Do not include any markdown, explanations, or additional text.\textbackslash n
\{src\_lang\} document: \{input\_text\}
\end{quote}

\paragraph{Rewriting Prompts}
Following \citet{rewriting}, we adopted three rewriting prompts:

\textit{Simplification:}
\begin{quote}\ttfamily
Simplify the \{src\_lang\} document. Simplification may include identifying complex words and replacing with simpler or shorter words or using active voice instead of passive voice. Try to keep the meaning of the Original document.\textbackslash n
Original: \{input\_text\}\textbackslash n
Simplified:
\end{quote}

\textit{Paraphrase:}
\begin{quote}\ttfamily
Paraphrase the \{src\_lang\} document. Try to not directly copy but keep the meaning of the Original document.\textbackslash n
Original: \{input\_text\}\textbackslash n
Paraphrase:
\end{quote}

\textit{Easy Translate:}
\begin{quote}\ttfamily
Rewrite the Original document to be easier for translation in \{tgt\_lang\} language. New document should be in \{src\_lang\}.\textbackslash n
Original: \{input\_text\}\textbackslash n
New:
\end{quote}

\section{Combining RLSR with QE-Guided Inference-Time Selection}
\label{sec:qe_selective}
\citet{rewriting_qe} proposed using a quality-estimation (QE) model to iteratively guide source rewriting at inference time.
Because their code is not publicly available and the paper does not provide sufficient implementation details to support faithful reproduction, we do not compare against their full method.
Instead, we examine whether RLSR can be combined with the core idea behind their approach: using a reference-free QE signal at inference time to decide whether a rewrite should be used.

We consider a simple variant, which we call \emph{selective rewriting}.
Given a source $s$, an RLSR model first generates a rewrite $\tilde{s}$.
The downstream MT model then produces translations for both the original and rewritten sources, $M(s)$ and $M(\tilde{s})$.
We use a reference-free QE model $Q_{\mathrm{QE}}$ to score both translations against the original source $s$, and choose the source that yields the higher QE score:
\begin{equation}
s_{\mathrm{sel}} =
\begin{cases}
\tilde{s}, & \text{if } Q_{\mathrm{QE}}(s, M(\tilde{s})) > Q_{\mathrm{QE}}(s, M(s)), \\
s, & \text{otherwise.}
\end{cases}
\end{equation}
The final translation is then $M(s_{\mathrm{sel}})$.
This procedure is more costly than standard RLSR because both the original and rewritten sources must be translated and scored.

\paragraph{Experimental Setup}
We used \texttt{xCOMET-XL} in QE mode as $Q_{\mathrm{QE}}$.
We applied selective rewriting to the RLSR models trained for Gemma3 27B and Qwen3 30B, and evaluated the resulting translations on the same WMT2025 General MT Shared Task test set and with the same evaluation metrics as in the main experiments.
\begin{table}[h]
\centering
\setlength\tabcolsep{1.5pt}
\small
\begin{tabular}{llcccccc}
\toprule
MT Model & Training & xCOMET & MetricX & GEMBA \\
\midrule
\multirow{3}{*}{Gemma3} & -  & 49.22 & 7.19 & 8.73 \\
& RLSR & 50.04& 6.99& 7.36\\
& RLSR (selective) & \textbf{50.66}&\textbf{6.96} & \textbf{7.34}\\
\midrule
\multirow{3}{*}{Qwen3} & - & 41.25 & 9.62& 15.62\\
& RLSR & 42.21 & 9.21 &\textbf{13.92}\\
& RLSR (selective) & \textbf{43.00}& \textbf{9.20} & 13.97\\
\bottomrule
\end{tabular}
\caption{Effect of QE-guided selective rewriting. We use \texttt{xCOMET-XL} in QE mode as the reference-free selector and apply it to RLSR models trained for Gemma3 27B and Qwen3 30B. Higher is better for xCOMET, while lower is better for MetricX and GEMBA.}
\label{tab:qe_select_exp}
\end{table}

\paragraph{Results}
Table~\ref{tab:qe_select_exp} shows that selective rewriting improves xCOMET and MetricX for both downstream MT models.
However, the gains on GEMBA are modest: selective rewriting improves GEMBA only slightly for Gemma3 27B and slightly worsens it for Qwen3 30B.
This pattern suggests that QE-guided selection can complement RLSR when evaluated by learned metrics, but its effect on the more independent GEMBA metric is limited.
We therefore view selective rewriting as a lightweight optional extension to RLSR.

\section{Experimental Results Grouped by Source Language}
\label{sec:src_lang_group_exp}

Owing to space constraints, the main text reports only the scores averaged over all 16 language pairs of the WMT2025 General MT Shared Task.
However, because the effectiveness of source rewriting may vary with the source language, we provide the full evaluation results grouped by source language.
The language pairs in the test set span three source languages: English, Czech, and Japanese. We group the results accordingly.
Tables~\ref{tab:main_exp_src_lang_gemma3}--\ref{tab:main_exp_src_lang_qwensftrl} present the per-source-language results for all six downstream MT models.

As these tables show, although statistical significance is not reached in a few isolated cases, RLSR still significantly outperforms the baselines for most metrics and MT models.
We therefore conclude that RLSR's performance gains are robust across source languages.

\begin{table*}[t]
\centering
\setlength\tabcolsep{1.5pt}
\small
\begin{tabular}{llccccccccc}
\toprule
\multirow{2}{*}{Rewriting Model} & \multirow{2}{*}{Rewriting Prompt} & \multicolumn{9}{c}{Source Language} \\
\cmidrule(lr){3-11}
 & & xCOMET & MetricX & GEMBA & xCOMET& MetricX & GEMBA & xCOMET& MetricX & GEMBA \\
\midrule
& & \multicolumn{3}{c}{English} & \multicolumn{3}{c}{Czech}& \multicolumn{3}{c}{Japanese}  \\
\midrule
- & - & 47.37 & 7.60 & 9.32 & 58.83&5.28&5.92&54.24&5.62& 6.61\\
\hdashline
Qwen3 4B & Simplification & 47.42 & 7.30 & 9.65 & 51.57 & 6.19 & 8.19 & 43.07 & 6.66 &8.54\\
Qwen3 4B & Paraphrase & 43.01 & 7.40 &  8.86 & 49.67 & 6.09 & 7.29 & 43.27 & 6.44 &7.59\\
Qwen3 4B & Easy Translate & 48.19 & 7.33 & 8.70 & 52.36 & 6.18 & 7.38 & 50.45 & 5.98 &7.13\\
\hdashline
Qwen3 235B & Simplification & 48.34 & 7.13 & 8.04 & 57.36 & 5.54 & 6.25 & 55.54 & 5.33 & 5.93\\
Qwen3 235B & Paraphrase & 42.42 & 7.39 & 9.70 & 52.90 & 5.35 & 7.02 & 53.63 & 5.40 & 7.15\\
Qwen3 235B & Easy Translate & \textbf{48.61} & \textbf{6.97} &\textbf{7.13} & 57.78 & 5.15 & 5.27 & \textbf{56.46} &\textbf{5.16} &\textbf{5.23}\\
\hdashline
Qwen3 4B (RLSR) & Easy Translate & ~~48.52\textsuperscript{†} & ~~7.11\textsuperscript{†} & ~~7.25\textsuperscript{†}& ~~\textbf{58.91}\textsuperscript{†}& ~~\textbf{5.11}\textsuperscript{†} & ~~5.04\textsuperscript{†} & ~~54.39\textsuperscript{†} & 5.61& ~~5.44\textsuperscript{†}\\
\rev{Qwen3 4B (RLSR-Ens)} & Easy Translate & ~~48.31\textsuperscript{†} & ~~7.41\textsuperscript{†} & ~~7.26\textsuperscript{†} & 58.82 & ~~5.12\textsuperscript{†} & ~~\textbf{4.97}\textsuperscript{†} & 54.31 & ~~5.52\textsuperscript{†} & ~~5.40\textsuperscript{†}\\
\bottomrule
\end{tabular}
\caption{Comparison of rewriting methods on the WMT2025 General MT Shared Task. The metric scores are grouped by source language. The MT model is \textbf{Gemma3 27B}. We performed a paired bootstrap test comparing each of Qwen3 4B (RLSR) and \rev{Qwen3 4B (RLSR-Ens)} against both the Qwen3 4B prompt-based rewriting baselines and the no-rewriting baseline; † indicates significantly better performance than all those baselines ($p < 0.05$).} 
\label{tab:main_exp_src_lang_gemma3}
\end{table*}

\begin{table*}[t]
\centering
\setlength\tabcolsep{1.5pt}
\small
\begin{tabular}{llccccccccc}
\toprule
\multirow{2}{*}{Rewriting Model} & \multirow{2}{*}{Rewriting Prompt} & \multicolumn{9}{c}{Source Language} \\
\cmidrule(lr){3-11}
 & & xCOMET & MetricX & GEMBA & xCOMET& MetricX & GEMBA & xCOMET& MetricX & GEMBA \\
\midrule
& & \multicolumn{3}{c}{English} & \multicolumn{3}{c}{Czech}& \multicolumn{3}{c}{Japanese}  \\
\midrule
- & - & 49.62 & 7.31 & 6.39 & 61.13 & 4.74 & 4.12 & 60.25 & 4.83 &4.31\\
\hdashline
Qwen3 4B & Simplification & 48.81 & 7.18 & 6.67& 52.61 & 5.97 & 5.55& 45.43 & 6.36 & 6.01\\
Qwen3 4B & Paraphrase & 43.81 & 7.29 & 7.33& 51.03 & 5.84 & 5.87& 46.10 & 6.01 & 6.15\\
Qwen3 4B & Easy Translate & 49.42 & 7.22 & 6.75& 54.63 & 5.77 & 5.60& 55.84 & 5.35 &5.18\\
\hdashline
Qwen3 235B & Simplification & 49.50 & 6.99 & 6.13& 59.59 & 5.22 & 4.55& 60.33 & 4.85 & 4.23\\
Qwen3 235B & Paraphrase & 43.40 & 7.29 & 8.21& 54.88 & 5.16 & 5.81& 56.85 & 4.99 &5.62\\
Qwen3 235B & Easy Translate & 49.39 & \textbf{6.92} & 6.04& 59.46 & 5.02 & 4.38& \textbf{61.71} & 4.74 &4.10\\
\hdashline
Qwen3 4B (RLSR) & Easy Translate & ~~\textbf{50.43}\textsuperscript{†} & ~~7.07\textsuperscript{†} & ~~\textbf{5.99}\textsuperscript{†}& ~~\textbf{61.93}\textsuperscript{†} & \textbf{4.70} & \textbf{4.02}& 60.26 & \textbf{4.72} & ~~4.11\textsuperscript{†}\\
\rev{Qwen3 4B (RLSR-Ens)} & Easy Translate & ~~50.32\textsuperscript{†} & ~~7.14\textsuperscript{†} & ~~6.08\textsuperscript{†} & ~~61.27\textsuperscript{†} & 4.72 & 4.05 & ~~60.40\textsuperscript{†} & 4.84 & ~~\textbf{4.00}\textsuperscript{†}\\
\bottomrule
\end{tabular}
\caption{Comparison of rewriting methods on the WMT2025 General MT Shared Task. The metric scores are grouped by source language. The MT model is \textbf{Gemma4 31B}. We performed a paired bootstrap test comparing each of Qwen3 4B (RLSR) and \rev{Qwen3 4B (RLSR-Ens)} against both the Qwen3 4B prompt-based rewriting baselines and the no-rewriting baseline; † indicates significantly better performance than all those baselines ($p < 0.05$).}
\label{tab:main_exp_src_lang_gemma4}
\end{table*}

\begin{table*}[t]
\centering
\setlength\tabcolsep{1.5pt}
\small
\begin{tabular}{llccccccccc}
\toprule
\multirow{2}{*}{Rewriting Model} & \multirow{2}{*}{Rewriting Prompt} & \multicolumn{9}{c}{Source Language} \\
\cmidrule(lr){3-11}
 & & xCOMET & MetricX & GEMBA & xCOMET& MetricX & GEMBA & xCOMET& MetricX & GEMBA \\
\midrule
& & \multicolumn{3}{c}{English} & \multicolumn{3}{c}{Czech}& \multicolumn{3}{c}{Japanese}  \\
\midrule
- & - & 56.39 & 5.28 & 5.79 & 62.32 & 4.33 & 4.75 & 61.60 & 4.39 & 4.92\\
\hdashline
Qwen3 4B & Simplification & 53.39 & 5.75 & 5.81 & 51.95 & 5.89 & 5.95 & 43.94 & 6.13 & 6.30 \\
Qwen3 4B & Paraphrase & 48.05 & 5.85 & 6.59 & 50.93 & 5.74 & 6.42 & 44.05 & 5.85 &  6.58\\
Qwen3 4B & Easy Translate& 54.70 & 5.49 & 5.82 & 55.27 & 5.39 &5.72  & 56.28 & 5.06 &  5.47\\
\hdashline
Qwen3 235B & Simplification & 54.73 & 5.50 &  6.08& 59.39 & 4.96 & 5.48 & 59.72 & 4.44 & 5.08 \\
Qwen3 235B & Paraphrase & 47.67 & 5.74 & 6.68 & 54.45 & 5.01 & 6.47 & 57.46 & 4.65 & 5.53 \\
Qwen3 235B & Easy Translate & 54.32 & 5.50 & 5.82 & 59.09 & 4.75 & 5.03 & 61.28 & 4.38 & 4.43 \\
\hdashline
Qwen3 4B (RLSR) & Easy Translate & ~~\textbf{58.10}\textsuperscript{†} & ~~\textbf{5.22}\textsuperscript{†} & ~~5.53\textsuperscript{†} & ~~\textbf{63.03}\textsuperscript{†} & \textbf{4.28} & ~~4.50\textsuperscript{†} & ~~\textbf{63.28}\textsuperscript{†} & \textbf{4.32} & ~~4.47\textsuperscript{†}\\
\rev{Qwen3 4B (RLSR-Ens)} & Easy Translate & ~~56.80\textsuperscript{†} & 5.27 & ~~\textbf{5.47}\textsuperscript{†} & ~~62.81\textsuperscript{†} & 4.30 & ~~\textbf{4.45}\textsuperscript{†} & 61.91 & 4.34 & ~~\textbf{4.40}\textsuperscript{†}\\
\bottomrule
\end{tabular}
\caption{Comparison of rewriting methods on the WMT2025 General MT Shared Task. The metric scores are grouped by source language. The MT model is \textbf{Translategemma 27B}. We performed a paired bootstrap test comparing each of Qwen3 4B (RLSR) and \rev{Qwen3 4B (RLSR-Ens)} against both the Qwen3 4B prompt-based rewriting baselines and the no-rewriting baseline; † indicates significantly better performance than all those baselines ($p < 0.05$).}
\label{tab:main_exp_src_lang_transgemma}
\end{table*}

\begin{table*}[t]
\centering
\setlength\tabcolsep{1.5pt}
\small
\begin{tabular}{llccccccccc}
\toprule
\multirow{2}{*}{Rewriting Model} & \multirow{2}{*}{Rewriting Prompt} & \multicolumn{9}{c}{Source Language} \\
\cmidrule(lr){3-11}
 & & xCOMET & MetricX & GEMBA & xCOMET& MetricX & GEMBA & xCOMET& MetricX & GEMBA \\
\midrule
& & \multicolumn{3}{c}{English} & \multicolumn{3}{c}{Czech}& \multicolumn{3}{c}{Japanese}  \\
\midrule
- & - & 38.22 & 10.51 & 16.69 & 52.97& 5.03&6.38&57.28&5.38&5.87 \\
\hdashline
Qwen3 4B & Simplification & 38.77 & 9.84 & 16.03 & 47.38 & 6.92 & 11.27 & 44.45 & 6.42 &10.62\\
Qwen3 4B & Paraphrase & 35.26 & 9.97 & 17.10 & 46.00 & 6.81 & 11.68 & 44.74 & 6.07 & 10.49\\
Qwen3 4B & Easy Translate & 39.16 & 9.72 & 15.57 & 48.11 & 7.03 & 11.26 & 53.92 & 5.60 & 9.15\\
\hdashline
Qwen3 235B & Simplification & 39.29 & 9.70 & 15.22 & 52.69 & 6.25 & 9.81 & 58.45 & \textbf{4.96} & 7.67\\
Qwen3 235B & Paraphrase & 34.98 & 9.99 & 16.83 & 48.76 & 6.04 & 10.17 & 55.77 & 5.06 &8.73\\
Qwen3 235B & Easy Translate & \textbf{39.57} & \textbf{9.57} & 14.90 & 53.18 & 6.07 & 8.82 & \textbf{59.54} & 5.04 & 7.13\\
\hdashline
Qwen3 4B (RLSR) & Easy Translate &  ~~39.15\textsuperscript{†} & ~~9.59\textsuperscript{†} & ~~\textbf{14.71}\textsuperscript{†} & ~~\textbf{53.45}\textsuperscript{†} & \textbf{5.01} & ~~\textbf{5.92}\textsuperscript{†} & ~~58.85\textsuperscript{†} & ~~5.12\textsuperscript{†} & ~~\textbf{5.40}\textsuperscript{†}\\
\rev{Qwen3 4B (RLSR-Ens)} & Easy Translate & ~~39.11\textsuperscript{†} & ~~10.15\textsuperscript{†} & ~~14.80\textsuperscript{†} & ~~53.18\textsuperscript{†} & 5.03 & ~~5.98\textsuperscript{†} & ~~58.88\textsuperscript{†} & ~~5.07\textsuperscript{†} & ~~5.44\textsuperscript{†}\\
\bottomrule
\end{tabular}
\caption{Comparison of rewriting methods on the WMT2025 General MT Shared Task. The metric scores are grouped by source language. The MT model is \textbf{Qwen3 30B}. We performed a paired bootstrap test comparing each of Qwen3 4B (RLSR) and \rev{Qwen3 4B (RLSR-Ens)} against both the Qwen3 4B prompt-based rewriting baselines and the no-rewriting baseline; † indicates significantly better performance than all those baselines ($p < 0.05$).}
\label{tab:main_exp_src_lang_qwen30b}
\end{table*}

\begin{table*}[t]
\centering
\setlength\tabcolsep{1.5pt}
\small
\begin{tabular}{llccccccccc}
\toprule
\multirow{2}{*}{Rewriting Model} & \multirow{2}{*}{Rewriting Prompt} & \multicolumn{9}{c}{Source Language} \\
\cmidrule(lr){3-11}
 & & xCOMET & MetricX & GEMBA & xCOMET& MetricX & GEMBA & xCOMET& MetricX & GEMBA \\
\midrule
& & \multicolumn{3}{c}{English} & \multicolumn{3}{c}{Czech}& \multicolumn{3}{c}{Japanese}  \\
\midrule
- & - & 30.18 & 13.34 & 23.07 & 40.30 & 9.47 & 16.37 & 53.25 & 5.62 & 9.20\\
\hdashline
Qwen3 4B & Simplification & 29.14 & 15.17 & 23.38 & 38.82 & 10.70 & 16.59 & 51.30 & 6.35 & 9.94 \\
Qwen3 4B & Paraphrase & 28.56 & 16.12 & 23.67 & 38.03 & 11.48 & 16.89 & 50.22 & 6.75 & 9.95 \\
Qwen3 4B & Easy Translate & 30.14 & 14.02 & 22.10 & 40.19 & 9.99 & 16.47 & 53.10 & 5.94 & 9.44  \\
\hdashline
Qwen3 235B & Simplification & 30.66 & 12.88 & 20.96 & 40.93 & 9.14 & 14.71 & 54.09 & 5.42 & 9.19\\
Qwen3 235B & Paraphrase & 29.50 & 15.21 & 22.21 & 39.39 & 10.79 & 16.40 & 52.05 & 6.40 &  9.68 \\
Qwen3 235B & Easy Translate & 31.00 & 12.35 & 20.43 & 41.39 & 8.53 & \textbf{13.43} & 54.69 & 5.20 & \textbf{8.42}  \\
\hdashline
Qwen3 4B (RLSR) & Easy Translate& ~~31.23\textsuperscript{†} & ~~\textbf{12.03}\textsuperscript{†} & ~~\textbf{20.23}\textsuperscript{†} & ~~\textbf{41.69}\textsuperscript{†} & ~~8.76\textsuperscript{†} &  ~~13.67\textsuperscript{†}& ~~55.10\textsuperscript{†} & ~~\textbf{5.06}\textsuperscript{†} & ~~8.63\textsuperscript{†}  \\
\rev{Qwen3 4B (RLSR-Ens)} & Easy Translate & ~~\textbf{31.46}\textsuperscript{†} & ~~12.23\textsuperscript{†} & ~~20.31\textsuperscript{†} & ~~41.19\textsuperscript{†} & ~~\textbf{8.42}\textsuperscript{†} & ~~13.88\textsuperscript{†} & ~~\textbf{55.19}\textsuperscript{†} & ~~5.11\textsuperscript{†} & ~~8.57\textsuperscript{†}\\
\bottomrule
\end{tabular}
\caption{Comparison of rewriting methods on the WMT2025 General MT Shared Task. The metric scores are grouped by source language. The MT model is \textbf{Qwen3 4B}. We performed a paired bootstrap test comparing each of Qwen3 4B (RLSR) and \rev{Qwen3 4B (RLSR-Ens)} against both the Qwen3 4B prompt-based rewriting baselines and the no-rewriting baseline; † indicates significantly better performance than all those baselines ($p < 0.05$).}
\label{tab:main_exp_src_lang_qwen4b}
\end{table*}

\begin{table*}[t]
\centering
\setlength\tabcolsep{1.5pt}
\small
\begin{tabular}{llccccccccc}
\toprule
\multirow{2}{*}{Rewriting Model} & \multirow{2}{*}{Rewriting Prompt} & \multicolumn{9}{c}{Source Language} \\
\cmidrule(lr){3-11}
 & & xCOMET & MetricX & GEMBA & xCOMET& MetricX & GEMBA & xCOMET& MetricX & GEMBA \\
\midrule
& & \multicolumn{3}{c}{English} & \multicolumn{3}{c}{Czech}& \multicolumn{3}{c}{Japanese}  \\
\midrule
- & - & 33.35 & 11.81 & 19.24 & 44.40 & 8.18 & 13.19 & 55.60 & 5.41 & 9.04 \\
\hdashline
Qwen3 4B & Simplification & 33.07 & 11.53 & 19.20 & 41.78 & 8.24 & 14.22  & 41.82 & 6.47 & 10.10 \\
Qwen3 4B & Paraphrase & 30.95 & 11.45 & 19.82 & 41.03 & 8.06 & 14.78 & 42.60 & 6.14 & 10.68 \\
Qwen3 4B & Easy Translate & 33.80 & 11.43 & 19.10 & 42.01 & 8.64 &  14.77 & 52.97 & 5.78 &  10.01 \\
\hdashline
Qwen3 235B & Simplification & 33.60 & 11.33 & 19.10 & 45.11 & 7.98 & 13.50 & 56.46 & 5.19 & 8.58 \\
Qwen3 235B & Paraphrase &30.96 & 11.46 & 19.08 & 42.52 & 7.66 & 12.25 & 54.05 & 5.16 & 8.76 \\
Qwen3 235B & Easy Translate & \textbf{33.82} & 11.16 & \textbf{17.28} & \textbf{46.07} & \textbf{7.52} & \textbf{11.22} & \textbf{57.93} & \textbf{4.93} & \textbf{7.72} \\
\hdashline
Qwen3 4B (RLSR) & Easy Translate & ~~33.62\textsuperscript{†} & ~~\textbf{11.05}\textsuperscript{†} & ~~17.44\textsuperscript{†} & ~~44.70\textsuperscript{†} & ~~7.63\textsuperscript{†} & ~~11.57\textsuperscript{†} & ~~56.01\textsuperscript{†} & ~~5.06\textsuperscript{†} & ~~8.11\textsuperscript{†} \\
\rev{Qwen3 4B (RLSR-Ens)} & Easy Translate & ~~33.68\textsuperscript{†} & ~~11.12\textsuperscript{†} & ~~17.66\textsuperscript{†} & ~~44.64\textsuperscript{†} & ~~7.89\textsuperscript{†} & ~~11.69\textsuperscript{†} & ~~56.13\textsuperscript{†} & ~~5.19\textsuperscript{†} & ~~8.41\textsuperscript{†} \\
\bottomrule
\end{tabular}
\caption{Comparison of rewriting methods on the WMT2025 General MT Shared Task. The metric scores are grouped by source language. The MT model is \textbf{Qwen3 4B SFT-RL}. We performed a paired bootstrap test comparing each of Qwen3 4B (RLSR) and \rev{Qwen3 4B (RLSR-Ens)} against both the Qwen3 4B prompt-based rewriting baselines and the no-rewriting baseline; † indicates significantly better performance than all those baselines ($p < 0.05$).}
\label{tab:main_exp_src_lang_qwensftrl}
\end{table*}

\section{Implementation Details of Training Rewriting Models via SFT}
\label{sec:sft}
The source--reference pairs used for SFT are the same as those used for RLSR (Section~\ref{sec:exp_setup}).
To construct a supervised target for each source, we prompted Qwen3 4B with the ``Easy Translate'' prompt (Appendix~\ref{sec:prompt}) to generate $N = 16$ distinct rewriting candidates.
For each downstream MT model, we then selected the candidate $\tilde{s}^*$ that yielded the largest \texttt{xCOMET-XL} improvement over the original source (i.e., the candidate with the highest reward $\mathcal{R}$ defined in Eq.~\ref{eq:reward}); when no candidate improved upon the original source, we used the original source itself as the target ($\tilde{s}^* = s$).
We fine-tuned Qwen3 4B on the resulting $(s, \tilde{s}^*)$ pairs for at most three epochs with an effective batch size of 512.
We used a learning rate of $2 \times 10^{-5}$ with a cosine decay scheduler and a warmup ratio of 0.05, and adopted the 8-bit AdamW optimizer.
The ``SFT (filtered)'' variant in Table~\ref{tab:sft_exp} uses the same configuration but excludes all unchanged pairs ($\tilde{s}^* = s$) from the training set.

\section{Training Rewriting Models via DPO}
\label{sec:dpo}
\paragraph{Motivation}
As discussed in Section~\ref{sec:why_rl}, the advantage of RL over SFT for training rewriting models may arise because RL forces the policy to discriminate between rewrites that are nearly identical in surface form yet receive markedly different rewards.
The Direct Preference Optimization (DPO) objective \cite{dpo} shares this property: it trains the model to separate a preferred (chosen) output from a rejected one, and such pairs typically differ only subtly.
Motivated by this similarity, we investigate whether DPO can serve as an effective alternative for training rewriting models.

\paragraph{Method}
For each input, DPO requires a pair consisting of a preferred output and a rejected output.
We obtain such pairs by reusing the candidate-generation procedure of our SFT setup (Appendix~\ref{sec:sft}): for each source $s$, we sample $N$ rewriting candidates $\{\tilde{s}_1, \dots, \tilde{s}_N\}$ using $R_{\text{ref}}$ and form $\mathcal{C}(s)=\{s, \tilde{s}_1, \dots, \tilde{s}_N\}$.
We score every candidate in $\mathcal{C}(s)$ with the reward $\mathcal{R}$ defined in Eq.~\ref{eq:reward}; thus, the original source is included as a no-rewriting candidate with reward $\mathcal{R}(s,s,r)=0$.
We then take the highest-reward candidate as the preferred output $\tilde{s}^{+}$ and the lowest-reward candidate as the rejected output $\tilde{s}^{-}$:
\begin{equation}
\begin{aligned}
    \tilde{s}^{+} &\in \operatorname*{arg\,max}_{\hat{s} \in \mathcal{C}(s)} \mathcal{R}(s, \hat{s}, r), \\
    \tilde{s}^{-} &\in \operatorname*{arg\,min}_{\hat{s} \in \mathcal{C}(s)} \mathcal{R}(s, \hat{s}, r).
\end{aligned}
\end{equation}
Under this construction, the unmodified source can serve as the preferred output when all generated rewrites are worse, or as the rejected output when all generated rewrites improve over it.
Given the resulting preference dataset $\mathcal{D}_{\text{pref}} = \{(s, \tilde{s}^{+}, \tilde{s}^{-})\}$, we train the rewriting model with the DPO objective:
\begin{equation}
\begin{aligned}
    \mathcal{L}_{\text{DPO}} = - \mathbb{E}_{\mathcal{D}_{\text{pref}}} \Big[ \log \sigma \big( & \beta \log \frac{R_{\theta}(\tilde{s}^{+}|s)}{R_{\text{ref}}(\tilde{s}^{+}|s)} \\
    & - \beta \log \frac{R_{\theta}(\tilde{s}^{-}|s)}{R_{\text{ref}}(\tilde{s}^{-}|s)} \big) \Big],
\end{aligned}
\end{equation}
where $\sigma(\cdot)$ is the logistic function, $R_{\theta}$ is the policy being trained, $R_{\text{ref}}$ is the frozen reference policy, and $\beta$ controls the strength of the regularization toward $R_{\text{ref}}$.
The sequence-level probabilities $R_{\theta}(\tilde{s}|s)$ and $R_{\text{ref}}(\tilde{s}|s)$ are defined as in Section~\ref{sec:rlsr}.

\paragraph{Experimental Setup}
We evaluated DPO with Gemma3 27B, Qwen3 30B, and Translategemma 27B as the downstream MT models.
The protocol for constructing preference pairs, including the reward model (\texttt{xCOMET-XL}) and the number of rewriting candidates per source ($N = 16$), is identical to that of SFT (Appendix~\ref{sec:sft}).
We set the KL divergence coefficient $\beta$ to $0.1$ and trained for at most three epochs with a learning rate of $1 \times 10^{-6}$, an effective batch size of 256, and the 8-bit AdamW optimizer.
We do not report SFT results for Translategemma 27B: because SFT already performed poorly on the other two MT models in the early stages of our study, we omitted this run to conserve computational resources.
\begin{table*}[h]
\centering
\setlength\tabcolsep{5pt}
\small
\begin{tabular}{llcccccc}
\toprule
MT Model & Training & xCOMET$\uparrow$ & MetricX$\downarrow$ & GEMBA$\downarrow$ \\
\midrule
\multirow{5}{*}{Gemma3} & -  & 49.22 & 7.19 & 8.73 \\
& RLSR & \textbf{50.04}& \textbf{6.99}& \textbf{7.36}\\
& SFT & 48.41&7.89 & 9.05\\
& SFT (filtered) & 46.82&9.66 & 10.63\\
& DPO & 49.41&7.27 & 9.21\\
\midrule
\multirow{5}{*}{Qwen3} & - & 41.25 & 9.62& 15.62\\
& RLSR & \textbf{42.21} &\textbf{9.21} &\textbf{13.92}\\
& SFT & 41.19&9.66 & 15.88\\
& SFT (filtered) & 40.95&9.72 & 16.01\\
& DPO & 42.03&9.32 & 14.97\\
\midrule
\multirow{3}{*}{Translategemma} & - & 57.45 & 5.11 & 5.97\\
& RLSR & \textbf{57.90} &\textbf{5.04} &\textbf{5.69}\\
& DPO & 53.30&6.60 & 9.51\\
\bottomrule
\end{tabular}
\caption{Performance comparison of rewriting models trained via RLSR, SFT, and DPO. ``Gemma3'', ``Qwen3'', and ``Translategemma'' denote Gemma3 27B, Qwen3 30B, and Translategemma 27B, respectively. ``SFT (filtered)'' removes instances with unchanged targets ($\tilde{s}^* = s$) from the SFT training set. SFT results for Translategemma 27B are omitted (see text).}
\label{tab:dpo_exp}
\end{table*}

\paragraph{Results}
Table~\ref{tab:dpo_exp} reports the results. Two findings stand out.
First, DPO largely outperforms both SFT variants on the two MT models for which SFT results are available. On Gemma3, DPO attains 49.41 xCOMET, compared with 48.41 for SFT and 46.82 for SFT (filtered); the same ordering holds on MetricX, although DPO trails SFT slightly on GEMBA (9.21 vs.\ 9.05). On Qwen3, DPO outperforms both SFT variants on all three metrics.
This is consistent with our hypothesis that an objective requiring the model to discriminate between similar rewrites of differing quality is better suited to this task than the plain maximum-likelihood objective of SFT.
Second, and more importantly, DPO is markedly less stable than RLSR.
RLSR improves over the no-rewriting baseline for all three MT models and all three metrics, whereas DPO does not. DPO yields clear gains on Qwen3, is roughly on par with the no-rewriting baseline on Gemma3 (a small xCOMET gain but slightly worse MetricX and GEMBA), and fails on Translategemma 27B, falling well below the no-rewriting baseline on every metric.
In every case, DPO trails RLSR by a clear margin.

\paragraph{Summary}
Overall, DPO provides a more effective training signal than SFT but a substantially less reliable one than RLSR.
We attribute this gap to the distinction between off-policy and on-policy learning.
RLSR is on-policy: it repeatedly samples rewrites from the current policy and obtains fresh reward feedback, allowing it to explore the rewriting space and correct its errors throughout training.
DPO, by contrast, is trained on a fixed set of preference pairs collected once from the reference model. This static dataset may cover the rewriting space only sparsely, leaving the policy unable to discover better rewriting strategies that on-policy exploration can reach. In the worst case (Translategemma 27B), it can even steer the model toward rewrites that actively harm translation quality.

\section{Computing Source Similarity}
\label{sec:src_sim_details}

The ``Src.\ sim.'' column of Table~\ref{tab:edit_locality} measures how much of the source a rewrite preserves. We use the character-level similarity ratio from Python's \texttt{difflib.SequenceMatcher}, i.e., $2M/(|s|+|\tilde{s}|)$, where $|s|$ and $|\tilde{s}|$ are the character lengths of the source $s$ and the rewrite $\tilde{s}$, and $M$ is the total number of characters in their matched blocks. These blocks are identified by recursively finding the longest common contiguous substring in the left and right remainders, following Ratcliff--Obershelp / gestalt matching. Figure~\ref{lst:srcsim} gives a self-contained implementation and a worked example: for the same source, the localized RLSR rewrite scores $0.667$ ($M{=}34$, $|s|{=}46$, $|\tilde{s}|{=}56$), whereas the wholesale Paraphrase rewrite scores only $0.468$.
\begin{figure*}[t]
\centering
\begin{lstlisting}[language=Python,
    basicstyle=\ttfamily\footnotesize,
    keywordstyle=\bfseries,
    commentstyle=\itshape,
    stringstyle=\itshape,
    showstringspaces=false,
    columns=fullflexible,
    breaklines=true,
    frame=single,
    framesep=5pt,
    xleftmargin=4pt, xrightmargin=4pt]
import re
from difflib import SequenceMatcher

def normalize(text):
    # In the data, source text stores newlines as the two literal
    # characters "\n", while rewrites use real newlines; we normalize both forms.
    text = text.replace("\\n", " ").replace("\n", " ")
    return re.sub(r"\s+", " ", text).strip()

def src_sim(source, rewrite):
    a, b = normalize(source), normalize(rewrite)
    # SequenceMatcher.ratio() == 2 * M / (len(a) + len(b)),
    # where M is the number of matched characters (gestalt matching).
    return SequenceMatcher(None, a, b).ratio()

source     = "Trump's bullshit blitz has Europe on its knees"
rlsr       = "Trump's barrage of nonsense has left Europe on its knees"
paraphrase = "Trump's relentless barrage of falsehoods has left Europe reeling."

print(src_sim(source, rlsr))        # 0.667   (M=34, |s|=46, |s~|=56)
print(src_sim(source, paraphrase))  # 0.468
\end{lstlisting}
\caption{Minimal implementation of the source-similarity metric used in Table~\ref{tab:edit_locality}, with a worked example. The localized RLSR edit retains far more of the source ($0.667$) than the wholesale Paraphrase rewrite ($0.468$).}
\label{lst:srcsim}
\end{figure*}

\rev{\section{Future Work}
\label{sec:future_work}
We see three promising directions for future work.

First, RLSR is considerably more expensive to train than SFT, so developing more efficient training methods could substantially reduce this cost. One concrete avenue, building on our analysis in Section~\ref{sec:why_rl}, is to improve the SFT objective, for example by up-weighting the few critical tokens that drive reward improvements \cite{token_sft}, so that it approaches the effectiveness of RLSR at a fraction of the training cost.

Second, source rewriting and translation could be unified within a single model through multi-task learning, in which one model is trained both to rewrite the source and to translate it. Beyond simplifying the pipeline, this raises the question of whether jointly learning to rewrite sources strengthens the model's own understanding of the translation task.

Third, the era of reasoning models invites a fresh look at the roles of source rewriting (pre-editing) and post-editing in MT. Both operations decompose translation into interpretable intermediate steps, which aligns naturally with the explicit reasoning chains of such models. For example, trained pre-editing models such as RLSR, together with post-editing models, could be used to construct chain-of-thought (CoT) training data for reasoning MT models: rewriting the source to remove translation obstacles, translating it, and then refining the output form a natural reasoning trajectory for the translation task.}

\end{document}